\documentclass[sigconf, nonacm]{acmart}

\newcommand\vldbpagestyle{empty} 

\usepackage{hyperref}
\usepackage{enumerate}
\usepackage{caption}
\usepackage{subcaption}
\usepackage{graphicx}
\usepackage[USenglish]{babel}
\usepackage{rotating}
\usepackage{tikz}
\usepackage{multicol}
\usepackage{multirow}
\usepackage{xcolor,colortbl}
\usepackage{algorithm}
\usepackage{textcomp}
\usetikzlibrary{backgrounds}
\usetikzlibrary{bayesnet}
\usetikzlibrary{arrows}

\usepackage{bbm}
\usepackage{bm}
\newtheorem{definition}{Definition}

\usepackage{array}
\usepackage{tabularx}
\usepackage{arydshln}
\usepackage{gensymb}
\newcolumntype{"}{@{\hskip\tabcolsep\vrule width 1pt\hskip\tabcolsep}}
\usepackage{soul}  %

\usepackage{makecell, multirow}
\setcellgapes{3pt}

\newboolean{show-tr}
\setboolean{show-tr}{true}  %
\newcommand{\techreport}[2]{\ifthenelse{\boolean{show-tr}}{{#1}}{{#2}}}

\newcommand{\divad}{DIVAD}
\newcommand{\divadg}{DIVAD-G}
\newcommand{\divadgm}{DIVAD-GM}
\newcommand{\smoothing}{\gamma}

\newcommand{\isep}{\mathrel{{.}\,{.}}\nobreak}

\newcommand\bsb[1]{\boldsymbol{#1}}

\definecolor{NavyBlue}{RGB}{8, 111, 189}  %
\definecolor{Red}{RGB}{255, 46, 23}  %
\definecolor{Green}{RGB}{0, 171, 79}  %

\newcommand{\skipzerolist}
    {\begin{list}{\hfil}
        {\topsep 0pt plus 1pt
           \parsep 1pt plus 1pt
           \partopsep 0pt plus 1pt
           \itemsep 0pt plus1pt
           \itemindent -0.2in }
    } 

\newcommand{\skipzeroitemize}
    {
      \begin{list}{{$\blacktriangleright$}}
        {
          \topsep 0pt plus 1pt
           \parsep 1pt plus 1pt
           \partopsep 0pt plus 1pt
           \itemsep 0pt plus 1pt 
           \itemindent -0.2in}
    }

\begin{document}
\title{Unsupervised Anomaly Detection in Multivariate Time Series across Heterogeneous Domains}

\author{Vincent Jacob}
\affiliation{%
  \institution{Ecole Polytechnique}
  \city{Palaiseau}
  \country{France}
}
\email{vincent.jacob@polytechnique.edu}

\author{Yanlei Diao}
\affiliation{%
  \institution{Ecole Polytechnique}
  \city{Palaiseau}
  \country{France}
}
\email{yanlei.diao@polytechnique.edu}

\begin{abstract}
The widespread adoption of digital services, along with the scale and complexity at which they operate, has made incidents in IT operations increasingly more likely, diverse, and impactful. This has led to the rapid development of a central aspect of ``Artificial Intelligence for IT Operations" (AIOps), focusing on detecting anomalies in vast amounts of multivariate time series data generated by service entities.  In this paper, we begin by introducing a unifying framework for benchmarking unsupervised anomaly detection (AD) methods, and highlight the problem of shifts in normal behaviors that can occur in practical AIOps scenarios. To tackle anomaly detection under domain shift, we then cast the problem in the framework of domain generalization and propose a novel approach, Domain-Invariant VAE for Anomaly Detection (\divad), to learn domain-invariant representations for unsupervised anomaly detection. Our evaluation results using the Exathlon benchmark show that the two main \divad\ variants significantly outperform the best unsupervised AD method in maximum performance, with 20\% and 15\% improvements in maximum peak F1-scores, respectively. Evaluation using the Application Server Dataset further demonstrates the broader applicability of our domain generalization methods.
\end{abstract}

\maketitle

\pagestyle{\vldbpagestyle}

\section{Introduction}
\label{sec:intro}

Time series anomaly detection has been studied intensively due to its broad application to domains such as financial market analysis, system diagnosis, and mechanical systems~\cite{tsad-survey,dlad-survey}. 
Recently, it has been increasingly adopted in an emerging domain known as ``Artificial Intelligence for IT operations" (AIOps)~\cite{gartner-2018}, which proposes to use AI to automate and optimize large-scale IT operations~\cite{zhong-2023}. Not long ago, the role of IT was to support the business. Today as digital services and applications become the primary way that enterprises serve and interact with customers, IT \textit{is} the business -- almost every business depends on the continuous performance and innovation of its digital services. 

With this paradigm shift, incidents in IT operations have become more impactful, inducing ever-increasing financial costs, both directly through service-level agreements made with customers and indirectly through brand image deterioration. Concurrently, the popularity of such services, along with their widespread migration to the cloud, has greatly increased the scale and complexity at which they operate, relying on more resources to process larger volumes of data at high speed. This evolution has made incidents more frequent, costly, diverse, and difficult for engineers to manually anticipate and diagnose, thus calling for more automated solutions.

To respond to such needs, this paper focuses on \emph{anomaly detection in multivariate time series} that suits the challenges in AIOps. More specifically, a large set of multivariate time series are generated from the periodic monitoring of service entities, and ``anomalies'' are reported as patterns in data that deviate from a given notion of \emph{normal behavior}~\cite{ad-survey}.

\textbf{Challenges.} 
Detecting anomalies in AIOps presents a set of technical challenges~\cite{zhong-2023}. (CH1) The \emph{scarcity of anomaly labels} is due to the lack of domain knowledge of IT operations to reliably label anomalies, and the labor-intensive process of examining large amounts of time series data. (CH2) The \emph{high dimensionality of recorded time series}, both in terms of time and feature dimensions, is common in AIOps due to the collection of numerous metrics at high frequency across a large number of entities. (CH3) The \emph{complexity and variety in normal behaviors} arise because multiple, complex entities are monitored at scale in different contexts. (CH4) The \emph{shifts in normal behaviors} further arise due to potentially frequent changes in software/service components, hardware components, or operation contexts of the monitored entities. 
The recent Exathlon benchmark~\cite{exathlon} exhibits significant shifts in normal behaviors across traces collected from different runs of Spark streaming applications. Similarly, the Application Server Dataset (ASD)~\cite{interfusion} exhibits shifts in behaviors of different servers.
In these cases, the shifts in normal behaviors are so significant that they appear to be samples collected from different \emph{domains} or \emph{contexts}. 

A large number of anomaly detection (AD) methods for multivariate time series have been developed, as categorized recently by Schmidl et al.~\cite{tsad-eval}. CH1 has been typically addressed through the development of \emph{unsupervised} AD methods, assuming no label information for training, and \emph{semi-supervised} methods that assume (possibly noisy) labels for the normal class only~\cite{ad-survey}. In this paper, we jointly refer to them as ``unsupervised'' methods, trained on mostly-normal data and evaluated on a labeled test set.
Concurrently, the advent of \emph{deep learning} (DL)~\cite{deep-learning} has been instrumental in partly addressing CH2 and CH3, offering the ability to learn succinct yet effective representations of high-dimensional data while capturing both temporal (i.e., intra-feature) and spatial (i.e., inter-feature) dependencies in multivariate time series~\cite{interfusion, mscred, tranad}. 
Despite covering a wide range of assumptions about both normal data and anomalies, all these methods are vulnerable to CH4, by assuming a similar distribution of training and test normal data, which makes them of limited use in the new AIOps scenario. 

This paper tackles the last challenge (CH4), compounded by other challenges (CH1-Ch3), through the framework of \textbf{domain generalization} (DG). In this framework, data samples are collected from multiple, distinct \emph{domains} (the normal contexts here), with certain characteristics of the observed data being determined by the domain, and others being independent from it. This amounts to associating shifts in normal behavior to the concept of \emph{domain shift}, and aiming to build models from a set of training (or \emph{source}) domains that can generalize to another set of test (or \emph{target}) domains. Most of existing DG methods were proposed for image classification, categorized as based on \emph{explicit feature alignment}, \emph{domain-adversarial learning} or \emph{feature disentanglement}~\cite{dg-survey, dg-survey-2}. In practice, adversarial methods can suffer from instabilities that make them hard to reproduce~\cite{gan-instability-1, gan-instability-2}, and explicit feature alignment become very costly as the number of source domains increases, like in AIOps. For these reasons, this paper focuses on feature disentanglement~\cite{diva, feature-disent-1, feature-disent-2}, where methods seek to decompose the input data into \emph{domain-shared} and \emph{domain-invariant} features. 
The existing methods,  designed for image classification, are not applicable to unsupervised time series anomaly detection (with no labels). Further, recent efforts on domain generalization for time series AD handle only univariate sound waves with various labeling assumptions~\cite{dcase2022-challenge}, making them unsuitable for the AIOps setting.

\textbf{Contributions.} 
In this paper, we present the first \emph{multivariate time series anomaly detection approach that generalizes across heterogeneous domains}. Given that this topic has been underaddressed in the anomaly detection literature, we conduct an in-depth study to characterize the problem of normal behavior shifts using the recent Exathlon benchmark~\cite{exathlon} (which was motivated by AIOps use cases) and to highlight the performance issues of existing unsupervised anomaly detection methods under domain shift. We then address this challenge by proposing a novel approach based on domain generalization and feature disentanglement, custom-designed for unsupervised time series anomaly detection. 
More specifically, our paper makes the following contributions: 

\begin{itemize}
    \item 
    We introduce a unifying framework for benchmarking unsupervised anomaly detection methods, and highlight the domain shift problem in AIOps scenarios (Section~\ref{sec:ad-analysis}). %
    
    \item 
    To tackle the problem of domain shift, we develop a theoretical formulation of unsupervised anomaly detection in the framework of \emph{domain generalization} (Section~\ref{sec:problem}). %
    
    \item 
    In this proposed framework, we develop a novel approach, called Domain-Invariant VAE for Anomaly Detection (\divad), with a set of variants to learn domain-invariant representations, thereby enabling effective anomaly detection in unseen domains (Section~\ref{sec:divad}).
\end{itemize}

Our evaluation using the Exathlon benchmark shows that our two main \divad\ variants can significantly outperform the best unsupervised AD method in maximum performance, with 20\% and 15\% improvements in maximum peak F1-scores ($0.79$ and $0.76$ over $0.66$), respectively. Our evaluation also applies \divad\ to the Application Server Dataset (ASD)~\cite{interfusion}, reflecting a similar use case, and shows that its explicit domain generalization can be more broadly applicable and useful in this second use case.

The code for our \divad\ method and experiments is available at \url{https://github.com/exathlonbenchmark/divad}.

\section{Related Work}
\label{sec:related}

\textbf{Anomaly Detection in Multivariate Time Series.} Numerous unsupervised anomaly detection methods in multivariate time series have been proposed over the years~\cite{ad-survey, dlad-survey, tsad-survey}. Schmidl et al.~\cite{tsad-eval} recently introduced a taxonomy based on the way the methods derive their \emph{anomaly scores} for data samples (the higher the score, the more deemed anomalous by the method). The only category we do not consider in this work is \emph{distance methods}, which typically do not scale well with the large dimensionality of AIOps. 

\emph{Forecasting methods} define anomaly scores of data samples as forecasting errors, based on the distance between the forecast and actual value(s) of one or multiple data point(s) in a context window of length $L$. LSTM-AD~\cite{lstm-ad} is the most popular forecasting method. It trains a stacked LSTM network to predict the next $l$ data records from the first $L-l$ of a window. It then fits a multivariate Gaussian distribution to the error vectors it produced in a validation set, and defines the anomaly score of a record as the negative log-likelihood of its error with respect to this distribution.

\emph{Reconstruction methods} score data samples based on their reconstruction errors from a transformed space. Principal Component Analysis (PCA)~\cite{outlier-book-ch3} and Autoencoder (AE)~\cite{replicator, ae} are representative shallow and deep reconstruction methods, respectively. PCA's transformation is a projection on the linear hyperplane formed by the principal components of the data, while AE's is a non-linear mapping to a latent encoding learned by a neural network that was trained to reconstruct data from it. More recently, Multi-Scale Convolutional Recurrent Encoder-Decoder (MSCRED)~\cite{mscred} turns a multivariate time series into multi-scale signature matrices characterizing system status at different time steps, and learns to reconstruct them using convolutional encoder-decoder and attention-based ConvLSTM networks. TranAD~\cite{tranad} relies on two transformer-based encoder-decoder networks, with the first encoder considering the current input window, and the second one considering a larger \emph{context} of past data in the window's sequence. It defines the anomaly score of an input window as the average of its reconstruction errors coming from two decoders and inference phases, with the second phase using the reconstruction error from the first phase as a focus score to detect anomalies at a finer level.

\emph{Encoding methods} score data samples based on their deviation within a transformed space. Deep SVDD~\cite{deep-svdd} is the most popular recent encoding method, training a neural network to map the input data to a latent representation enclosed in a small hypersphere, and defining anomaly scores of test samples as their squared distance from this hypershere's centroid. More recently, DCDetector~\cite{dcdetector} uses a dual-view attention structure based on contrastive learning to derive representations where differences between normal points and anomalies are amplified, subdividing windows into adjacent ``patches", with one view modeling relationships within patches and the other across patches. To derive anomaly scores, it uses the insight that normal points tend to be similarly correlated for both views, while anomalies tend to be more correlated to their adjacent points than to the rest of the window.

\emph{Distribution methods} define anomaly scores of data samples as their deviation from an estimated distribution of the data. The Mahalanobis method~\cite{pca, outlier-book-ch3} and Variational Autoencoder (VAE)~\cite{baseline-vae} are representative shallow and deep distribution methods, respectively. The Mahalanobis method estimates the data distribution as a multivariate Gaussian, and defines the anomaly score of a test vector as its squared Mahalanobis distance from it. VAE estimates it using a variational autoencoder, with the anomaly score of a test point derived by drawing multiple samples from its probabilistic encoder, and averaging the negative log-likelihood of the reconstructions obtained from each of these samples. A more recent method is OmniAnomaly~\cite{omni-anomaly}. It estimates the distribution of multivariate windows with a stochastic recurrent neural network, explicitly modeling temporal dependencies among variables through a combination of GRU and VAE. It then defines a test window's anomaly score as the negative log-likelihood of its reconstruction.

\emph{Isolation tree methods} score data samples based on their ``isolation level" from the rest of the data. Isolation forest~\cite{iforest} is the most popular isolation tree method. It trains an ensemble of trees to isolate the samples in the training data, and defines the anomaly score of a test instance as inversely proportional to the average path length required to reach it using the trees. 

Overall, these methods cover a wide range of assumptions about both normal data and anomalies. As we show in this paper, \emph{by assuming a similar distribution of training and test normal data, all of them are vulnerable to shifts in normal behavior}, limiting their applicability in our AIOps scenario.

\textbf{Domain Generalization.} Domain generalization (DG) has been mainly studied in the context of \emph{image classification}, with the domains usually corresponding to the way images are represented or drawn. DG methods can broadly be categorized as based on \emph{explicit feature alignment}, \emph{domain-adversarial learning} or \emph{feature disentanglement}~\cite{dg-survey, dg-survey-2}. Explicit feature alignment methods seek to learn data representations where feature distribution divergence is explicitly minimized across domains, with divergence metrics including the Wasserstein distance or Kullback-Leibler divergence~\cite{dg-survey, dg-survey-2}. Rather than using such divergence metrics directly, domain-adversarial learning methods seek to minimize domain distribution discrepancy through a minimax two-player game, where the goal is to make the features confuse a domain discriminator~\cite{dann}, usually implemented as a domain classifier~\cite{multiclass-adv-1, multiclass-adv-2, binary-adv-1, binary-adv-2}. Such adversarial methods can suffer from instabilities that make them hard to reproduce~\cite{gan-instability-1, gan-instability-2}, while explicit feature alignment can become very costly as the number of source domains increases, like in AIOps. For these reasons, this work considers domain generalization based on feature disentanglement~\cite{diva, feature-disent-1, feature-disent-2}, where methods seek to decompose the input data into \emph{domain-specific} and \emph{domain-invariant} features, and perform their tasks in domain-invariant space.

Our work is specifically related to Domain-Invariant Variational Autoencoders (DIVA)~\cite{diva}, designed for image classification. It uses variational autoencoders (VAE) to decompose input data into domain-specific, class-specific, and residual latent factors, conditioning the distributions of its domain-specific and class-specific factors on the training domain and class, respectively, and enforcing this conditioning by using classification heads to predict the domain and class from the corresponding embeddings. It then uses its class-related classifier to derive its predictions for the test images. Because of this class supervision, this method cannot be applied to our unsupervised AD setting.

Domain generalization for time series AD recently gained attention through anomalous sound detection and the DCASE2022 Challenge, where the task was to identify whether a machine was normal or anomalous using only normal sound data under domain-shifted conditions~\cite{dcase2022-challenge}. The methods proposed 
however modeled single-channel (univariate) sound waves, while also 
assuming labels such as the machine state, the type of machine, domain shift or noise considered to train domain-invariant or disentangled representations~\cite{sound-disentanglement}. 
This univariate aspect, coupled with these simplifying assumptions, makes such methods unsuitable for our AIOps setting.

\textbf{Data Drift Detection.}
Many techniques exist for data drift detection~\cite{gama2014survey}. However, popular methods such as using the Kolmogorov-Smirnov distance~\cite{dos2016fast,bu2017incremental} require a significant amount of drifted data to detect a distribution change accurately. Anomaly detection under domain shift is essentially a different problem, where anomalies must be detected with low latency as they arise, although the normal behaviors in the current domain may appear to be drawn from a different context from those seen in training data.

\section{General AD Framework}
\label{sec:ad-analysis}

In this section, we present the unsupervised anomaly detection (AD) problem, propose a unifying framework to encompass AD approaches in evaluation, and highlight the presence of domain shift in the current framework using the Exathlon~\cite{exathlon} benchmark.

\subsection{Unsupervised Anomaly Detection}
\label{subsec:ad-def}

We first introduce the notation of the paper and define the AD problem in the unsupervised setting. 
More specifically, we consider $N_1$ training sequences and $N_2$ test sequences:
$$
\mathcal S_{\text{train}} = (\bsb S^{(1)}, \ldots, \bsb S^{(N_1)}) \ , \ \mathcal S_{\text{test}} = (\bsb S^{(N_1 + 1)}, \ldots, \bsb S^{(N_1 + N_2)}),
$$
\noindent where each $\bsb S^{(i)}$ consists of $T$ ordered data records of dimension $M$. 
To simplify the notation, our problem definition uses $T$ to denote the (same) length of all sequences, while our techniques do not make this assumption and can handle variable-length sequences.  

For each \emph{test} sequence, we consider a sequence of \emph{anomaly labels}:
$$
\mathcal Y_{\text{test}} = \{\bsb y^{(N_1 + 1)}, \ldots, \bsb y^{(N_1 + N_2)}\},
$$
\noindent with $\bsb y^{(i)} \in \{0, 1\}^T$, such that:
\begin{equation*}
\begin{cases}
  y_t^{(i)} = 1 & \text{if the record at index } t \text{ in sequence } i \text{ is \emph{anomalous},} \\
  y_t^{(i)} = 0 & \text{otherwise (i.e., the record is \emph{normal}). }
\end{cases}
\end{equation*}

Our goal is to build an anomaly detection model as follows.

\begin{definition}
An anomaly detection model is a \emph{record scoring function} $g : \mathbb R^{T \times M} \rightarrow \mathbb R^T$, mapping a sequence $\bsb S$ to a sequence of real-valued record-wise anomaly scores $g(\bsb S)$, which assigns higher anomaly scores to anomalous records than to normal records in test sequences. That is, 
$g(\bsb S^{(i)})_{t_1} > g(\bsb S^{(j)})_{t_2}$, $\forall i, j \in [N_1 \isep N_1 + N_2]$,  $t_1, t_2 \in [1 \isep T]$ s.t.  $y^{(i)}_{t_1} = 1 \wedge y^{(j)}_{t_2} = 0$. 
	\end{definition}

This record scoring function should further be constructed in a setting of \emph{offline training} and \emph{online inference}. More precisely, it means that training has to be performed offline on $\mathcal S_{\text{train}}$, and inference must be performed online on $\mathcal S_{\text{test}}$ by  considering only the data  preceding a given record at time index $t$:
$$
g(\bsb S^{(i)})_t = g(\bsb S^{(i)}_{1:t})_t \ , \ \forall i \in [N_1 \isep N_1 + N_2] \ , \ t \in [1 \isep T]. 
$$
\noindent Due to this requirement, we refer to the anomaly detection methods based on $g$ as \emph{online scorers}.

\subsection{Unifying Anomaly Detection Framework}
\label{subsec:ad-framework}

We next propose a unifying framework to encompass AD approaches within a common evaluation structure. 
In this framework, each online scorer relies on a \emph{windowing operator} $W_L$ that extracts sliding windows, or \emph{samples}, of length $L > 0$ from a given sequence $i$: 
$$
W_L(\bsb S^{(i)}) = \{\bsb S_{t-L+1:t}^{(i)}\}_{t=L}^T =: \{\bsb x_t^{(i)}\}_{t=L}^T,
$$
\noindent with $\bsb x_t^{(i)} \in \mathbb R^{L \times M}$. 
Then the training set is composed of the samples extracted from all the training sequences:
$$
\mathcal D_{\text{train}} := \bigcup_{i \in [1 \isep N_1]} \left\{W_L(\bsb S^{(i)})\right\}.
$$

\begin{definition}
A \emph{window scorer}, trained on $D_{\text{train}}$, is a \emph{window scoring function} that assigns an anomaly score to a given window, proportional to its abnormality for the method, 
$g_W:$ $\mathbb R^{L \times M} \to \mathbb R$, 
$\bsb x \mapsto g_W(\bsb x)$.
\end{definition}

We encapsulate each individual AD method within a window scorer, and then propose a universal \emph{online scorer} constructed from the window scorer. 
Given a test sequence $\bsb S$ and a smoothing factor $\smoothing \in [0, 1)$, the online scorer assigns anomaly scores as follows:
\begin{equation*}
g(\bsb S; L, \smoothing)_t  = 
\begin{cases}
  -\infty & \text{if } t < L, \\
  (1 - \smoothing)\hat{y}_L =: m_L & \text{if } t = L, \\
  \frac{\smoothing m_{t-1} + (1 - \smoothing)\hat{y}_t}{(1 - \smoothing^{t+1})} =: m_t & \text{if } t > L.
\end{cases}
\end{equation*}
\noindent With:
$$
\hat{y}_t = g_W(\bsb S_{t-L+1:t}) \ , \ \forall t \in [L \isep T].
$$

\noindent In other words, for a test sequence, we assign an anomaly score to the current sliding window of length $L$ using $g_W$ (which is fixed and trained offline). We then define this window score as the anomaly score of its last record (i.e., the one just received in an online setting). To allow additional control on the tradeoff between the ``stability" and ``reactivity" of the record scoring function\footnote{This tradeoff is also influenced by the window length $L$, the window scoring function and the types of anomalies.}, we further apply an \emph{exponentially weighted moving average} with smoothing hyperparameter $\smoothing$ to the anomaly scores. This produces the final output of the record scoring function $g$ for a test sequence, prepended with infinitely low anomaly scores for the timestamps before its first full window of length $L$. In this framework, both $L$ and $\smoothing$ are hyperparameters to set for every anomaly detection method.

\begin{figure}
    \centering
    \begin{tabular}{lc}

	\hspace{-0.2in}
    \begin{subfigure}[b]{0.25\textwidth}
        \centering
        \includegraphics[width=\textwidth]{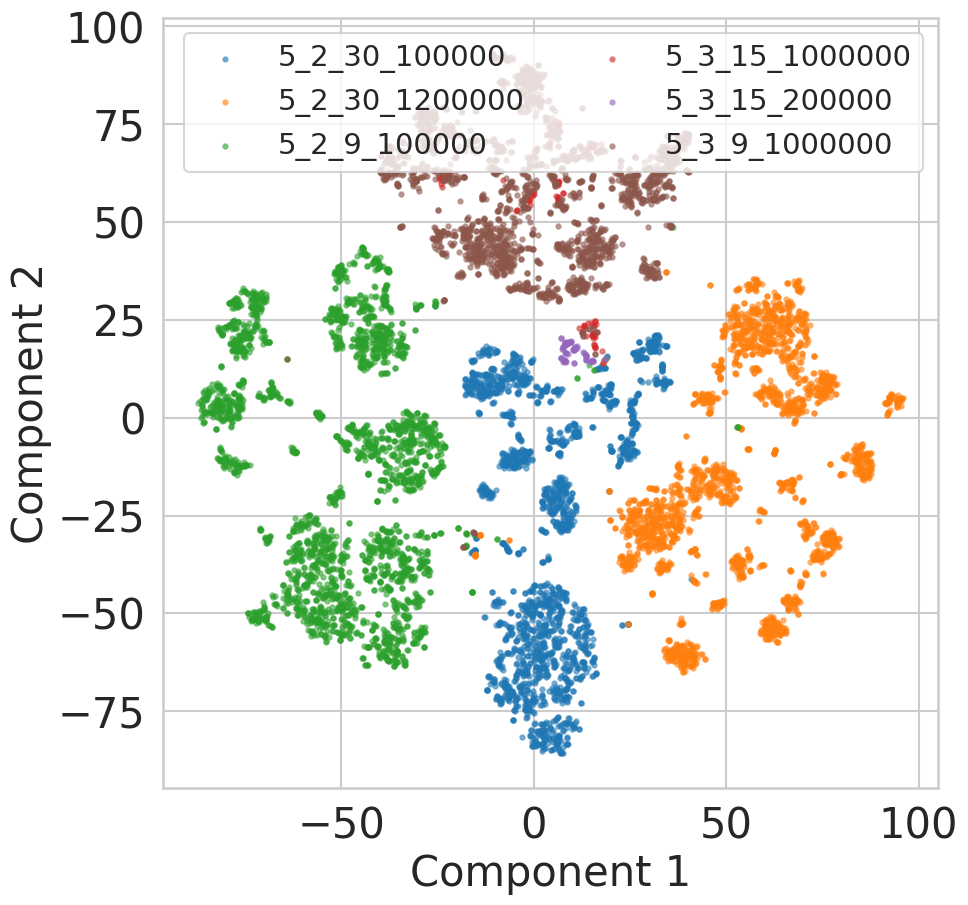}
        \caption{\small{Records colored by context.}}
        \label{fig:app2-diversity}
    \end{subfigure}

	&

    \begin{subfigure}[b]{0.25\textwidth}
        \centering
        \includegraphics[width=\textwidth]{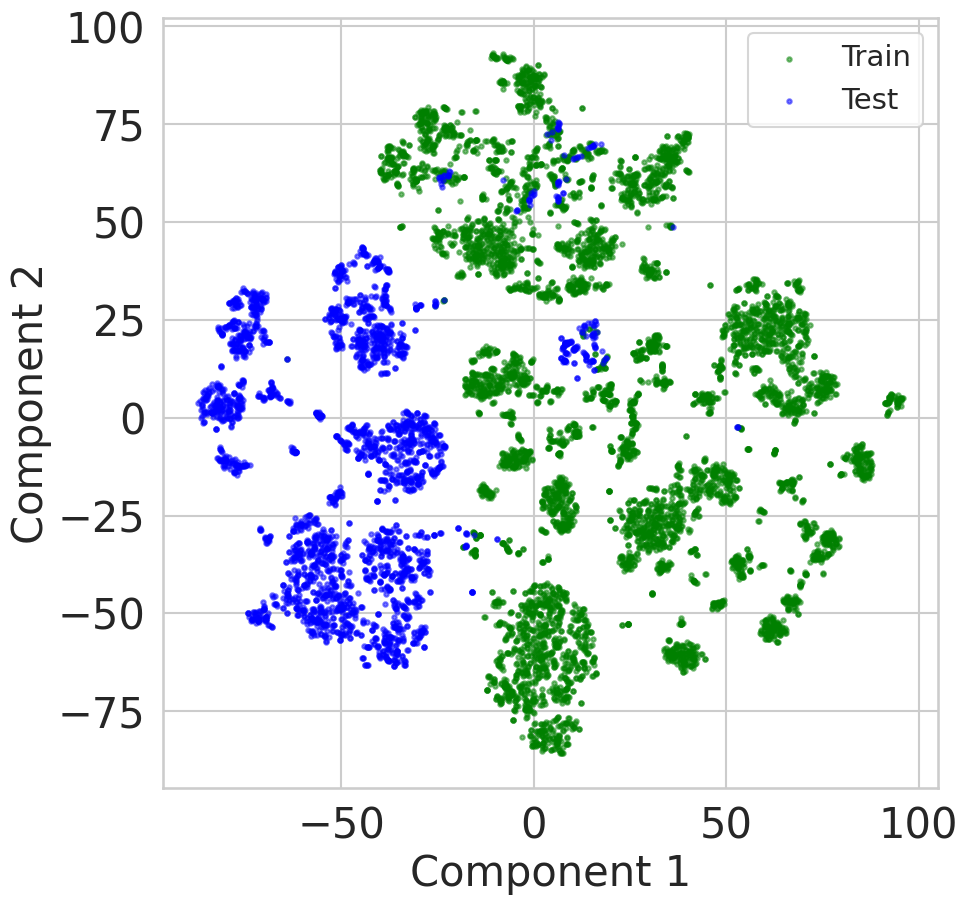}
        \caption{\small{Records colored by dataset.}}
        \label{fig:app2-shift}
    \end{subfigure}
      
    \end{tabular}
      
    \vspace{-0.2in}  
    \caption{\small{t-SNE scatter plots of application 2's normal data, undersampled to 10,000 data records balanced by context.}}
    \label{fig:app2-diversity-shift}
    \Description[\small t-SNE scatter plots of application 2's normal data, undersampled to 10,000 data records balanced by context]{t-SNE scatter plots of application 2's normal data, undersampled to 10,000 data records balanced by context}
    \vspace{-0.1in}
\end{figure}

\subsection{Domain Shift in Exathlon}

Inspired by real-world AIOps use cases, the Exathlon benchmark \cite{exathlon} is one of the most challenging benchmarks for anomaly detection due to the high-dimensionality and complex, diverse behaviors in its dataset, as reported in a recent comprehensive experimental study of AD methods~\cite{tsad-eval}. 

\subsubsection{Background on Exathlon}
Exathlon~\cite{exathlon} has been systematically constructed based on repeated executions of distributed Spark streaming \emph{applications} in a cluster under different Spark \emph{settings} and \emph{input rates}. The dataset includes 93 repeated executions of 10 Spark applications, with one \emph{trace} collected for each execution containing 2,283 raw features, resulting in a total size of 24.6GB. While 59 traces were collected in normal execution (\emph{normal traces}), 34 other traces were disturbed manually by injecting anomalous events (\emph{disturbed traces}). 
There are 6 different classes of anomalous events 
 (e.g., misbehaving inputs, resource contention, process failures), with a total of 97 anomalous instances.
For each of these anomalies, Exathlon provides the ground truth label for the interval spanning the root cause event and its lasting effect,  enabling accurate evaluation of AD methods. 
In addition to anomalous instances, both the normal and disturbed traces contain enough variety (e.g., Spark's checkpointing activities) to capture diverse normal behaviors.

\subsubsection{Analysis of Domain Shift}
\label{subsec:shift-example}

Given the limited number of Spark \emph{applications} (or \emph{entities}~\cite{aiops-ad}) in the Exathlon dataset, our study focuses on the ability of an AD method to generalize, not to new applications, but rather to the new \emph{contexts} of the Spark applications. 
For this reason, the \textbf{domain} of a trace is defined as its \emph{context}, characterized by the following factors:
(i) The Spark \emph{settings} for each application run includes its 
processing period (i.e., batch interval or window slide), set to a specific value for the application,  
the number of active executors and 
``memory profile" (i.e., maximum memory set for the driver block manager, executors JVM, and garbage collection). 
The last two aspects had either a direct or indirect impact on a lot of features (e.g., executors memory usage).
(ii) The \emph{input rate} is the rate at which data records were sent to the application, which had a direct effect on many recorded features (e.g., last completed batch processing delay).
The ``normal behavior'' in a trace is, therefore, mainly determined by its trace characteristics, defined as the combination of its \emph{entity} and \emph{domain/context}.

To illustrate the \emph{diversity} and \emph{shift} in domains/contexts, Figure~\ref{fig:app2-diversity-shift} shows t-SNE scatter plots~\cite{tsne} of application 2's normal data, undersampled to 10,000 data records. %
The \emph{diversity} is shown in Figure~\ref{fig:app2-diversity}, where data records are colored by context (where context labels include the processing period, number of Spark executors, maximum executors memory, and data input rate). We see that different contexts appear as distinct clusters, constituting a \emph{multimodal} distribution for data records. 
Figure~\ref{fig:app2-shift} illustrates the \emph{shift} in context:  
the different contexts induce a distribution shift from the training to the test data, even within normal records of the same application---we refer to this phenomenon as the \emph{domain shift} problem.

\section{Problem of Domain Shift} %
\label{sec:problem}

In this section, we formally define the problem of anomaly detection under domain shift. We took inspiration from one of the first adopted definitions of an \emph{anomaly} proposed by Douglas M. Hawkins in 1980, describing it as ``an observation which deviates so much from the other observations as to arouse suspicions that it was generated by a different mechanism"~\cite{hawkins}. 

This definition naturally suggests addressing our AD problem from a \emph{generative} perspective, assuming data samples were generated from a distribution $p_{\text{data}}(\mathbf x, \mathrm y)$\footnote{To simplify the notation, we use $y$ to refer to the label of an input \emph{sample} in the following: $y=0$ if the sample is (fully) normal, $y=1$ otherwise.}, with \emph{normal} samples generated from $p_{\text{data}}(\mathbf x \vert \mathrm y = 0)$. Our general goal then translates to constructing a {\bf model} $p_{\bsb \theta}(\mathbf x)$ of $p_{\text{data}}(\mathbf x \vert \mathrm y = 0)$, parameterized by $\bsb \theta \in \Theta$. This yields a natural definition for the {\bf anomaly score} of a test sample, as its negative log-likelihood with respect to this model:

$$
    g_W(\bsb x; \bsb \theta) := -\log p_{\bsb \theta}(\mathbf x = \bsb x)
$$
\noindent 
Since normal samples from the training and test sets are assumed generated from $p_{\text{data}}(\mathbf x \vert \mathrm y = 0) \approx p_{\bsb \theta}(\mathbf x)$, we would indeed expect them to have a higher likelihood under this model than anomalous samples, generated from $p_{\text{data}}(\mathbf x \vert \mathrm y = 1) \ne p_{\bsb \theta}(\mathbf x)$. 

A unique aspect of AIOps scenarios,  however, is that the distribution generating an observed sample can be conditioned not only on its class, but also on the specific \emph{sequence} this sample was extracted from. In particular, each sequence corresponds to a {\bf domain/context} that impacts the distribution of observed data, even for the same entity being recorded.  These domains can be included in our generative model, by assuming that the selection of a sequence $i$ corresponds to the realization $d_i$ of a discrete random domain variable  $\mathrm d \sim p_{\text{data}}(\mathrm d)$ with infinite support\footnote{We use $d_i$ (as opposed to $i$) to reflect the fact that multiple sequences can correspond to the same context, and thus value $d$ (i.e., we can have $d_i = d_j$ for $i \ne j$).}. In this setting, the samples of class $c \in \{0, 1\}$ from sequence $i$ 
 can be seen as independently drawn from a sequence-induced, or \emph{domain} distribution:
$$
p_i(\mathbf x \vert \mathrm y = c) = p_{\text{data}}(\mathbf x \vert \mathrm y = c, \mathrm d=d_i).
$$

\begin{figure}
    \centering
    \begin{tabular}{lc}
	\hspace{-0.1in}
    \begin{subfigure}[b]{0.2\textwidth}
        \centering
   	\includegraphics[width=0.6\columnwidth,height=1.8cm]{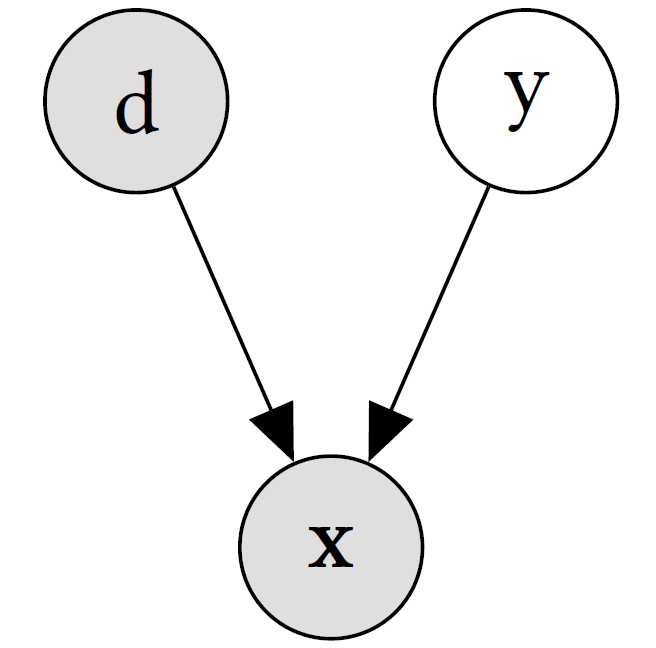}
        \caption{\small{Observed variable $\mathbf x$ depends on its domain $\mathrm d$ and latent (unobserved) class $\mathrm y$.}}
        \label{fig:gm-pb}
    \end{subfigure}
	&
    \begin{subfigure}[b]{0.26\textwidth}
        \centering
        \includegraphics[width=0.7\columnwidth,height=1.8cm]{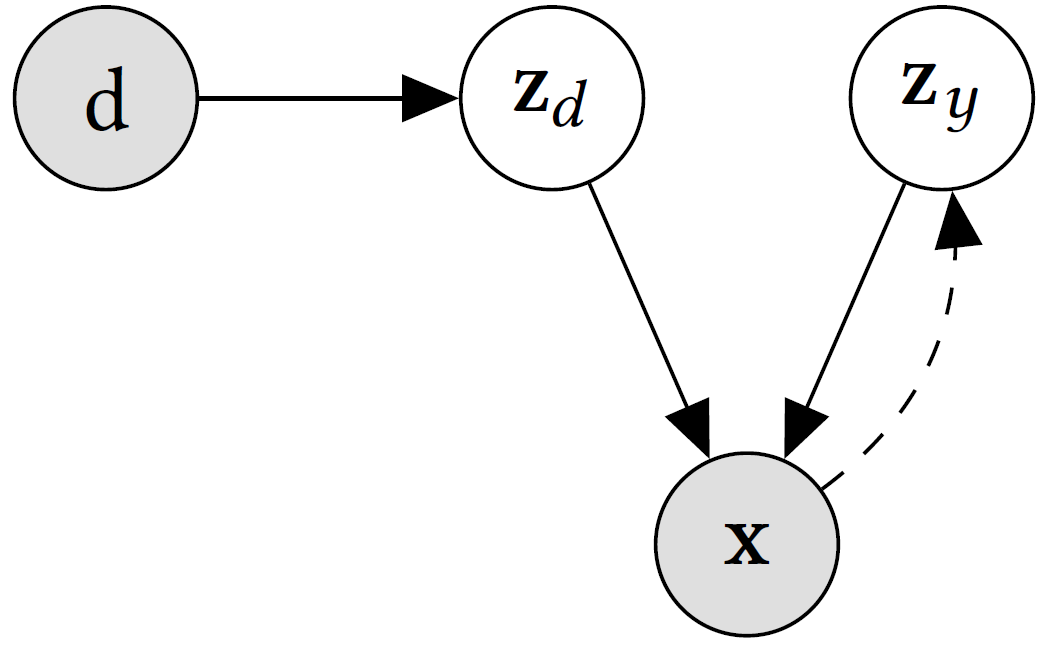}
    \caption{\small{$\mathbf x$ is caused by independent domain-specific $\mathbf z_d$ and domain-independent $\mathbf z_y$.}}
    \label{fig:gm-unsupervised}
    \end{subfigure}
    \end{tabular}
    \vspace{-0.15in}  
    \caption{\small{Generative models. For (b), constructing $f_y$ amounts to \emph{inferring} $\mathbf z_y$ from $\mathbf x$ (dashed arrow).}}
    \label{fig:generative-models}
    \Description[Generative models]{Generative models}
\end{figure}

This amounts to assuming the distribution of $\mathbf x$ is conditioned on the two independent variables $\mathrm d$ and $\mathrm y$, the former determining the domain the sample originates from, and the latter determining whether the sample is normal or anomalous. We illustrate the corresponding generative model in Figure~\ref{fig:gm-pb}. Under this model, the data-generating distribution of normal samples can be expressed as the countable mixture of all possible domain distributions:
$$
p_{\text{data}}(\mathbf x \vert \mathrm y = 0) = \sum_{d=1}^\infty p_{\text{data}}(\mathbf x \vert \mathrm y = 0, \mathrm d = d)p_{\text{data}}(\mathrm d=d).
$$

\begin{definition}[Domain Shift Challenge]
Directly applying traditional generative methods in an unsupervised setting amounts to making $p_{\bsb \theta}(\mathbf x)$ estimate the data-generating distribution of the normal \textbf{training} samples (with $d_i$'s fixed and all samples equally-likely to come from every sequence $i$):
$$
p_{\text{\emph{train}}}(\mathbf x \vert \mathrm y = 0) = \frac1 {N_1}\sum_{i=1}^{N_1} p_{\text{\emph{data}}}(\mathbf x \vert \mathrm y = 0, \mathrm d = d_i),
$$

\noindent which, given the infinitude of possible domains, is likely to differ from the data-generating distribution of the normal \textbf{test} samples:
$$
p_{\text{\emph{test}}}(\mathbf x \vert \mathrm y = 0) = \frac 1 {N_2}\sum_{i=N_1 + 1}^{N_1 + N_2} p_{\text{\emph{data}}}(\mathbf x \vert \mathrm y = 0, \mathrm d = d_i),
$$
\noindent with $\{d_i\}_{i=1}^{N_1} \neq \{d_i\}_{i=N_1 + 1}^{N_1 + N_2}$. This mismatch induces a \emph{domain shift challenge}, characterized by test normal samples $\bsb x_0 \sim p_{\text{\emph{test}}}(\mathbf x \vert \mathrm y = 0)$ and test anomalous samples $\bsb x_1 \sim p_{\text{\emph{test}}}(\mathbf x \vert \mathrm y = 1)$ being both unlikely in uncontrollable ways under $p_{\bsb \theta}(\mathbf x) \approx p_{\text{\emph{train}}}(\mathbf x \vert \mathrm y = 0)$, which hinders anomaly detection performance.
\end{definition}

A suitable framework to address this domain shift challenge is \emph{domain generalization}~\cite{dg-survey, dg-survey-2}. In this framework, the domains sampled for training are referred to as \emph{source domains}, while those sampled at test time are called \emph{target domains}.

\begin{definition}[Anomaly Detection with Domain Generalization]
\label{definition:ad-dg}
Our problem can be framed as \emph{building an AD model from the source domains that generalizes to the target domains}. We do so by assuming that the \emph{observed} variable $\mathbf x$ can be mapped via $f_y$ to a \emph{latent} representation $\mathbf z_y$, whose distribution is (i) discriminative with respect to the class $\mathrm y$ (i.e., normal vs. anomalous) and (ii) independent from the domain $\mathrm d$. Our goal can be formulated as:
\begin{itemize}
    \item Finding such a mapping $f_y(\mathbf x) = \mathbf z_y$; 
    \item Constructing $p_{\bsb \theta}(\mathbf x)$ to estimate $p_{\text{\emph{train}}}(f_y(\mathbf x) \vert \mathrm y = 0)$ instead of $p_{\text{\emph{train}}}(\mathbf x \vert \mathrm y = 0)$.
\end{itemize}

\end{definition}

Since $f_y(\mathbf x) = \mathbf z_y$ is independent from $\mathrm d$, we then have:
\begin{align*}
p_{\text{train}}(\mathbf z_y \vert \mathrm y = 0) &= \frac 1 {N_1} \sum_{i=1}^{N_1} p(\mathbf z_y \vert \mathrm y = 0, \mathrm d = d_i) \\
&= p(\mathbf z_y \vert \mathrm y = 0) = p_{\text{test}}(\mathbf z_y \vert \mathrm y = 0),
\end{align*}

\noindent which means that, under $p_{\bsb \theta}(\mathbf x) \approx p_{\text{train}}(\mathbf z_y \vert \mathrm y = 0) = p_{\text{test}}(\mathbf z_y \vert \mathrm y = 0)$, the \textbf{normal test samples} $\bsb x_0$ \textbf{should be more likely than the test anomalies} $\bsb x_1$, hence addressing the domain shift challenge.

\begin{figure}[t]
    \begin{center}
    \includegraphics[width=0.84\columnwidth,height=2.0cm]{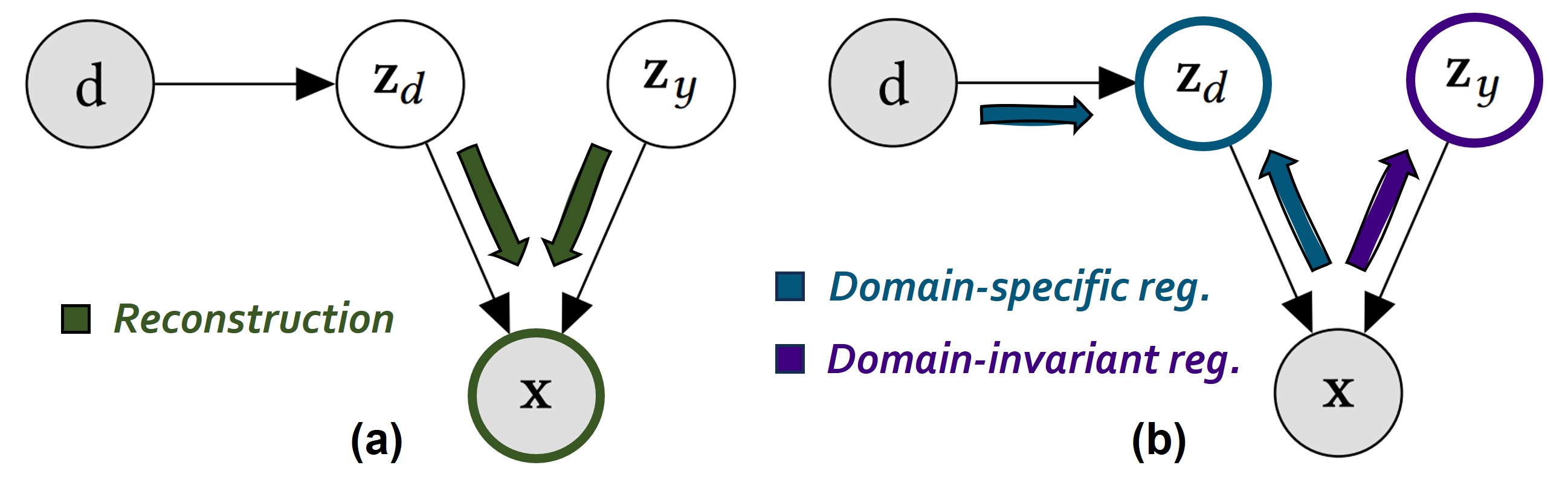}
    \vspace{-0.1in}
    \caption{\small Illustration of the (a) reconstruction and (b) regularization terms in the ELBO objective (Eq.~\ref{equation:elbo}).}
    \label{fig:gm-illustration}
    \Description[Illustration of the (a) reconstruction and (b) regularization terms in the ELBO objective (Eq.~\ref{equation:elbo}).]{Illustration of the (a) reconstruction and (b) regularization terms in the ELBO objective (Eq.~\ref{equation:elbo}).} 
    \end{center}
\end{figure}

\section{The \divad\ Method}
\label{sec:divad}

In this section, we introduce a new approach to anomaly detection under domain shift. 
At a high level, our central assumption is that \emph{anomalies should have a sensible impact on the properties of the input samples that are invariant with respect to the domain}. In an AIOps scenario, this means that, although some aspects of a running process may vary from domain to domain (e.g., its memory used or processing delay), others typically remain constant and characterize its ``normal" behavior (e.g., its scheduling delay, processing delay per input record, or any other Key Performance Indicator (KPI)~\cite{aiops-ad} that behaves similarly across contexts). These domain-invariant, normal-specific characteristics tend to reflect whether the process \emph{functions properly}, while domain-specific characteristics simply manifest \emph{different modes of normal operation}.

To realize this intuition, we propose Domain-Invariant VAE for Anomaly Detection, or \divad. This method embodies (i)~a new generative model with feature disentanglement to decompose the input data into domain-invariant and domain-specific factors, (ii)~an effective training approach in the VAE framework, with a custom training objective for our unique AD model, and (iii)~different alternatives for model inference, deriving anomaly scores based on the training distribution of domain-invariant factors only.

\subsection{Modeling}
\label{subsec:divad-model}
Based on the notion of \emph{feature disentanglement}~\cite{dg-survey, dg-survey-2}, our model assumes that the observed variable $\mathbf x$ is caused by two \emph{independent} latent factors $\mathbf z_d$ and $\mathbf z_y$: (i) $\mathbf z_d$ is conditioned on the observed domain $\mathrm d$, and (ii) $\mathbf z_y$ is assumed independent from it and can be used to detect anomalies in test samples. 
The corresponding generative model is shown in Figure~\ref{fig:gm-unsupervised}. 
Assuming that the model is parameterized by \emph{model parameters} $\bsb \theta \in \Theta$, the marginal likelihood $p_{\bsb \theta}(\mathbf x \vert \mathrm d)$ can be derived based on the structure of the generative model: 
\begin{align*}
    p_{\bsb \theta}(\mathbf x \vert \mathrm d) &= \int p_{\bsb \theta}(\mathbf x, \mathbf z_d, \mathbf z_y \vert \mathrm d) d \mathbf z_d d\mathbf z_y \\
    &= \int p_{\bsb \theta}(\mathbf x \vert \mathbf z_y, \mathbf z_d) p_{\bsb \theta}(\mathbf z_d \vert \mathrm d) p(\mathbf z_y) d \mathbf z_d d\mathbf z_y.
    \end{align*}

\subsection{Model Training}
\label{subsec:unsupervised-training}

We seek to learn the model parameters of $p_{\bsb \theta}(\mathbf x \vert \mathrm d)$ through maximum likelihood estimation. Since computing $p_{\bsb \theta}(\mathbf x \vert \mathrm d)$ directly is intractable, we leverage a variational autoencoder (\textbf{VAE}) \textbf{framework}~\cite{vae-2013, vae-2014}, considering \emph{variational parameters} $\bsb \phi_d, \bsb \phi_y \in \Phi$, and optimizing the following evidence lower bound (\textbf{ELBO}) instead, known to be a lower bound on $p_{\bsb \theta}(\mathbf x \vert \mathrm d)$ and lead to effective learning of its model parameters:
\begin{equation}
\begin{split}
\label{equation:elbo}
&\mathcal L_{\text{ELBO}}(\bsb x, d; \bsb \theta_{yd}, \bsb \theta_d, \bsb \phi_d, \bsb \phi_y) = \\
&\mathbb E_{q_{\bsb \phi_d}(\mathbf z_d \vert \bsb x)q_{\bsb \phi_y}(\mathbf z_y \vert \bsb x)}[\log p_{\bsb \theta_{yd}}(\bsb x \vert \mathbf z_d, \mathbf z_y)] \\ &- \beta D_{\text{KL}}(q_{\bsb \phi_y}(\mathbf z_y \vert \bsb x) \Vert p(\mathbf z_y)) - \beta D_{\text{KL}}(q_{\bsb \phi_d}(\mathbf z_d \vert \bsb x) \Vert p_{\bsb \theta_d}(\mathbf z_d \vert d)). 
\end{split}
\end{equation}
\noindent where the KL divergence terms are weighted by a factor $\beta$~\cite{beta-vae, diva}.

Figure~\ref{fig:gm-illustration} illustrates the effects of the different terms in the above training objective.
The first term of Eq.~\ref{equation:elbo} involves the \emph{likelihood} $p_{\bsb \theta_{yd}}(\bsb x \vert \mathbf z_d, \mathbf z_y)$. It measures \divad's ability to \emph{reconstruct} an input from its latent factors, $\mathbf z_d$ and $\mathbf z_y$, as shown in Figure~\ref{fig:gm-illustration}a.
The second and third terms act as domain-invariant and domain-specific \emph{regularizers}, pushing the \emph{variational posteriors} $q_{\bsb \phi_y}(\mathbf z_y \vert \bsb x)$ and $q_{\bsb \phi_d}(\mathbf z_d \vert \bsb x)$ toward their \emph{priors} $p(\mathbf z_y)$ and $p_{\bsb \theta_d}(\mathbf z_d \vert d)$, respectively, as illustrated by Figure~\ref{fig:gm-illustration}b. The prior $p(\mathbf z_y)$ will be used for anomaly scoring and further detailed in Section~\ref{subsec:anomaly-scoring}. The remaining distributions are learned using neural networks:
\begin{align*}
p_{\bsb \theta_{yd}}(\mathbf x \vert \mathbf z_d, \mathbf z_y) &= \mathcal N(\text{NN}_{\bsb \theta_{yd}}(\mathbf z_d, \mathbf z_y), \text{NN}_{\bsb \theta_{yd}}(\mathbf z_d, \mathbf z_y)) \\
p_{\bsb \theta_d}(\mathbf z_d \vert \mathrm d) &= \mathcal N(\text{NN}_{\bsb \theta_d}(\mathrm d), \text{NN}_{\bsb \theta_d}(\mathrm d)) \\
q_{\bsb \phi_y}(\mathbf z_y \vert \mathbf x) &= \mathcal N(\text{NN}_{\bsb \phi_y}(\mathbf x), \text{NN}_{\bsb \phi_y}(\mathbf x)) \\
q_{\bsb \phi_d}(\mathbf z_d \vert \mathbf x) &= \mathcal N(\text{NN}_{\bsb \phi_d}(\mathbf x), \text{NN}_{\bsb \phi_d}(\mathbf x)),
\end{align*}
\noindent 
where $\mathcal N$ denotes a Gaussian distribution with mean and variance each modeled by $\text{NN}_{\bsb \theta}(\cdot)$, a neural network with parameters $\bsb \theta$.
 
The conditional prior $p_{\bsb \theta_d}(\mathbf z_d \vert \mathrm d)$ has the effect of making $\mathbf z_d$ more dependent on $\mathrm d$, by ensuring that signals from $\mathrm d$ are incorporated into $\mathbf z_d$ (and thus facilitating the classification of $\mathrm d$ given $\mathbf z_d$). To further facilitate this domain classification, we add to maximum likelihood the following \textbf{domain classification} objective:
$$
\mathcal L_d(\bsb x, d; \bsb \phi_d, \bsb \omega_d) = \mathbb E_{q_{\bsb \phi_d}(\mathbf z_d \vert \bsb x)}\log q_{\bsb \omega_d}(d \vert \mathbf z_d),
$$
\noindent with $\bsb \omega_d \in \Omega$ the \emph{domain classifier parameters}. This objective amounts to training a domain classification head, by minimizing the cross-entropy loss based on the source domain labels. 

We perform gradient ascent on the final maximization objective:
\begin{multline*}
\mathcal L(\bsb x, d; \bsb \theta_{yd}, \bsb \theta_d, \bsb \phi_d, \bsb \phi_y, \bsb \omega_d) = \\\mathcal L_{\text{ELBO}}(\bsb x, d; \bsb \theta_{yd}, \bsb \theta_d, \bsb \phi_d, \bsb \phi_y) + \alpha_d \mathcal L_d(\bsb x, d; \bsb \phi_d, \bsb \omega_d), 
\end{multline*}
\noindent where $\alpha_d \in \mathbb R$ is a tradeoff hyperparameter balancing maximum likelihood estimation and domain classification. We do not share the parameters of our encoder networks $\text{NN}_{\bsb \phi_y}$ and $\text{NN}_{\bsb \phi_d}$, but instead consider a multi-encoder architecture.

\divad\ is similar in spirit to \emph{Domain-Invariant Variational Autoencoders} (DIVA)~\cite{diva}, proposed for image classification. However, our new problem setting of unsupervised anomaly detection leads to major differences from classification-based DIVA. First, by not relying on training class labels, \divad\ fuses DIVA's class-conditioned and residual latent factors $\mathbf z_y$ and $\mathbf z_x$ into a single, unconditioned domain-invariant factor $\mathbf z_y$, considering a conditioning and auxiliary classification objective only for the domain-specific factor $\mathbf z_d$. Second, rather than relying on an explicit classifier on top of the class-specific factor $\mathbf z_y$, \divad\ derives its anomaly scores from these factors' training distribution, modeled with the flexibility described in the following section.

\subsection{Model Inference}
\label{subsec:anomaly-scoring}

Based on Definition~\ref{definition:ad-dg}, our inference goals are to: (1) find a mapping $f_y(\mathbf x) = \mathbf z_y$ from the input to domain-invariant space, and (2) model the training distribution of $\mathbf z_y$ to derive our anomaly scores.

For task (1), a known result from VAE~\cite{vae-2013, vae-2014} is that after training based on Eq.~\ref{equation:elbo}, the variational posterior $q_{\phi_y}(\mathbf z_y \vert \mathbf x)$ should approximate the true posterior $p_{\bsb \theta}(\mathbf z_y \vert \mathbf x)$ (dashed arrow in Figure~\ref{fig:gm-unsupervised}). We can therefore use it to construct our mapping $f_y$:
$$
\mathbf z_y = f_y(\mathbf x) \sim q_{\bsb \phi_y}(\mathbf z_y \vert \mathbf x) \approx p_{\bsb \theta}(\mathbf z_y \vert \mathbf x).
$$

For task (2), we propose two alternatives below to model the training distribution of $\mathbf z_y$: prior and aggregated posterior estimate.

\subsubsection{Scoring from Prior}

\noindent In the first alternative, we derive the \textbf{anomaly score} $g_W(\bsb x)$ of a sample $\bsb x$ as the negative log-likelihood of $f_y(\bsb x)$ with respect to the prior $p(\mathbf z_y)$:
\begin{equation}
\label{equation:fixed-prior-scoring}
\boxed{
    g_W(\bsb x) := - \log p(\mathbf z_y = f_y(\bsb x))
}
\end{equation}

The rationale behind this method is the following: First, we observe that maximizing Eq.~\ref{equation:elbo} on average on the training set amounts to maximizing the regularization term of the ELBO w.r.t. $\mathbf z_y$: 
$$
\Omega_{\bsb \phi_y} := -\mathbb E_{p_{\text{train}}(\mathbf x)} D_{\text{KL}}(q_{\bsb \phi_y}(\mathbf z_y \vert \mathbf x) \Vert p(\mathbf z_y)). 
$$
\noindent However, based on~\cite{prior-blog}, we have the result:
\begin{multline*}
\Omega_{\bsb \phi_y} = \int \frac{1}{N_{\text{train}}}\sum_{i=1}^{N_{\text{train}}} q_{\bsb \phi_y}(\mathbf z_y \vert \bsb x_i) \log p(\mathbf z_y)d\mathbf z_y \\ - \int \frac{1}{N_{\text{train}}}\sum_{i=1}^{N_{\text{train}}} q_{\bsb \phi_y}(\mathbf z_y \vert \bsb x_i) \log q_{\bsb \phi_y}(\mathbf z_y \vert \bsb x_i)d\mathbf z_y,
\end{multline*}

\noindent where $N_{\text{train}}$ is the number of training samples. By considering:
$$
q_{\bsb \phi_y}(\mathbf z_y) = \frac{1}{N_{\text{train}}}\sum_{i=1}^{N_{\text{train}}} q_{\bsb \phi_y}(\mathbf z_y \vert \bsb x_i),
$$

\noindent the \textbf{marginal}, or \textbf{aggregated posterior}~\cite{adversarial-ae, fixing-elbo} (here the empirical distribution of encoded, presumably domain-invariant, samples), we therefore have:
\begin{align*}
\Omega_{\bsb \phi_y} &= \int q_{\bsb \phi_y}(\mathbf z_y) \log p(\mathbf z_y)d\mathbf z_y  \nonumber \\ &\quad - \int \frac{1}{N_{\text{train}}}\sum_{i=1}^{N_{\text{train}}} q_{\bsb \phi_y}(\mathbf z_y \vert \bsb x_i) \log q_{\bsb \phi_y}(\mathbf z_y \vert \bsb x_i)d\mathbf z_y \\
\Omega_{\bsb \phi_y} &= -H(q_{\bsb \phi_y}(\mathbf z_y), p(\mathbf z_y)) + H(q_{\bsb \phi_y}(\mathbf z_y \vert \mathbf x)),
\end{align*}

\noindent where $H(q_{\bsb \phi_y}(\mathbf z_y), p(\mathbf z_y))$ is the cross-entropy between the aggregated posterior and the prior, and $H(q_{\bsb \phi_y}(\mathbf z_y \vert \mathbf x))$ is the conditional entropy of $q_{\bsb \phi_y}(\mathbf z_y \vert \mathbf x)$ with the empirical distribution $\hat p_{\text{train}}(\mathbf x)$~\cite{prior-blog}. As we can see, \textbf{the maximization process of the ELBO has the effect of trying to make the aggregated posterior $q_{\bsb \phi_y}(\mathbf z_y)$ match the prior $p(\mathbf z_y)$}, which \textit{a priori} motivates the choice above of using the prior to derive anomaly scores.

\begin{figure*}
	\begin{center}
		\vspace{-0.1in}
		\includegraphics[width=1.9\columnwidth,height=7.9cm]{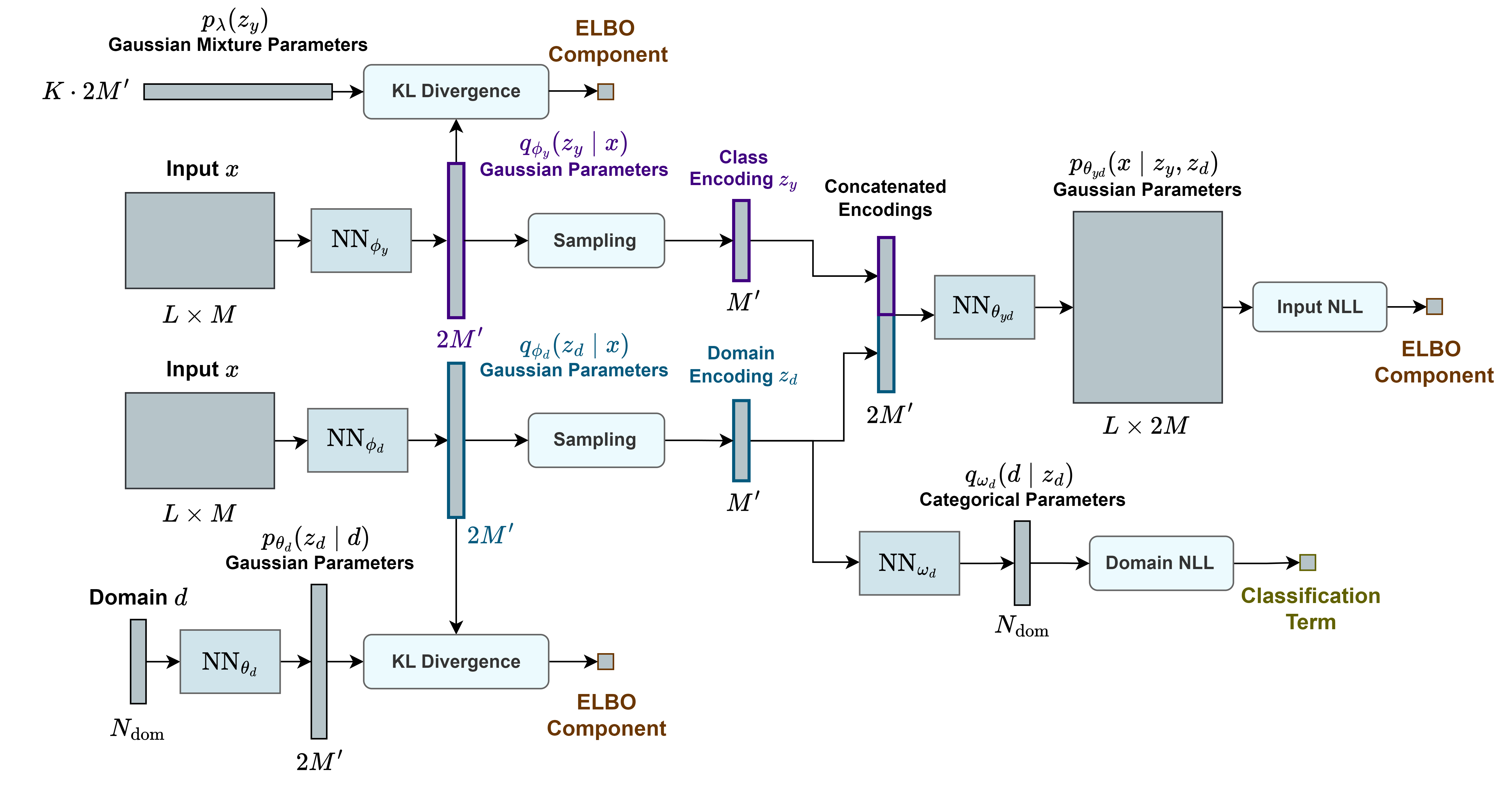}
		\vspace{-0.2in}
		\caption{\small Multi-encoder architecture of our \divadgm\ models, with $N_{\text{dom}}$ the number of source domains (\divadg\ models use a similar architecture, with the learned Gaussian Mixture parameters replaced with fixed Gaussian parameters).}
		\vspace{-0.1in}
		\label{fig:divad-multi-encoder}
		\Description[Multi-encoder architecture of our DIVAD models (DIVAD-G models use a similar architecture, with the learned Gaussian Mixture parameters replaced with fixed Gaussian parameters).]{Multi-encoder architecture of our DIVAD models (DIVAD-G models use a similar architecture, with the learned Gaussian Mixture parameters replaced with fixed Gaussian parameters).}
	\end{center}
\end{figure*}

\underline{Fixed Standard Gaussian Prior.} We first consider the default choice of prior for $\mathbf z_y$ in VAEs, as a fixed standard Gaussian:
$$
p(\mathbf z_y) = \mathcal N(\bsb 0, \bsb I),
$$

\noindent and refer to the method that uses this $\mathbf z_y$ prior and scores with Eq.~\ref{equation:fixed-prior-scoring} as \textbf{\divadg}. A limitation of \divadg~is that, although the aggregated posterior and prior \emph{should} be brought closer when maximizing the ELBO, they usually do not end up matching in practice at the end of training~\cite{fixing-elbo, dist-matching-vae}. This phenomenon is sometimes described as ``\emph{holes in the aggregated posterior}”, referring to the regions of the latent space that have high density under the prior but very low density under the aggregated posterior~\cite{resampled-priors}. 

\underline{Learned Gaussian Mixture Prior.} A method that has been shown to (at least partly) address the problem of aggregated posterior holes is to replace the fixed $\mathbf z_y$ prior with a \emph{learnable prior} $p_{\bsb \lambda}(\mathbf z_y)$~\cite{resampled-priors, prior-blog}, and hence have the maximization process update both the aggregated posterior and the prior. If sufficiently expressive, the prior can serve as a good approximation $\hat q_{\bsb \phi_y}(\mathbf z_y)$ of the aggregated posterior at the end of training, which makes it safer to use for anomaly scoring:
\begin{equation}
\label{equation:learned-prior-scoring}
\boxed{
    g_W(\bsb x) := - \log p_{\bsb \lambda}(\mathbf z_y = f_y(\bsb x))
}
\end{equation}

In a way, considering a learnable prior amounts to explicitly performing a joint density estimation of the marginal likelihood and aggregated posterior. With sufficiently expressive priors, this joint estimation also has the effect of \emph{putting less constraints on the aggregated posterior}, letting it capture normal clusters with more variance and arbitrary shapes. This is particularly useful in AD, where the ``normal" class can refer to a variety of different behaviors (even for the same entity). In practice, any density estimator $p_{\bsb \lambda}(\mathbf z_y)$ can be used to model the aggregated posterior. In this work, we consider a Gaussian Mixture (GM) distribution with $K$ components:
$$
p_{\bsb \lambda}(\mathbf z_y) = \sum_{k=1}^K w_k \mathcal N(\bsb \mu_k, \bsb \sigma_k^2),
$$

\noindent with $\bsb \lambda = \{w_k, \bsb \mu_k, \bsb \sigma_k^2\}_{k=1}^K$ randomly initialized and trained along with the other parameters. We refer to the method that uses this $\mathbf z_y$ prior and scores with Eq.~\ref{equation:learned-prior-scoring} as \textbf{\divadgm}.

\subsubsection{Scoring from Aggregated Posterior Estimate}
\label{subsec:agg-post-scoring}

An alternative (or complementary) solution to the problem of aggregated posterior holes is to perform the density estimation of the aggregated posterior $\hat q_{\bsb \phi_y}(\mathbf z_y)$ separately, and then define the anomaly score with respect to this estimate instead of the prior:
\begin{align}
\boxed{
    g_W(\bsb x) := - \log \hat q_{\phi_y}(\mathbf z_y = f_y(\bsb x)) 
}
\end{align}

In the following, we consider this alternative in addition to the prior-based scoring for both \divadg~and \divadgm. For \divadg, the aggregated posterior is estimated by fitting a multivariate Gaussian distribution to the training samples in latent space. For \divadgm, it is estimated by fitting to them a Gaussian Mixture model with the same number of components $K$ as the prior.

\subsection{Putting It All Together}
\label{subsec:all-together}

We illustrate the multi-encoder architecture of our \divad\ method in Figure~\ref{fig:divad-multi-encoder}, shown here for the learned Gaussian Mixture prior detailed in Section~\ref{subsec:anomaly-scoring}. From this figure, we can see that encoder networks $\text{NN}_{\bsb \phi_d}$ and $\text{NN}_{\bsb \phi_y}$ take the same sample $\bsb x$ as input to output the mean and variance parameters of multivariate Gaussians $q_{\bsb \phi_d}(\mathbf z_d \vert \mathbf x)$ and $q_{\bsb \phi_y}(\mathbf z_y \vert \mathbf x)$, respectively. These parameters are first used to compute the KL divergence terms of Equation~\ref{equation:elbo}, with the parameters of the conditional prior $p_{\bsb \theta_d}(\mathbf z_d \vert \mathrm d)$ outputted by a network $\text{NN}_{\bsb \theta_d}$ from the domain $d$ of $\bsb x$, and the parameters of $p_{\bsb \lambda}(\mathbf z_y)$ learned as described in Section~\ref{subsec:anomaly-scoring}. They are then used to sample the corresponding domain and class encodings of $\bsb x$: $\bsb z_d$ and $\bsb z_y$. These encodings, considered here of same dimension $M'$, are further concatenated to form the input of the decoder $\text{NN}_{\bsb \theta_{yd}}$, outputting the parameters of the multivariate Gaussian $p_{\bsb \theta_{yd}}(\mathbf x \vert \mathbf z_d, \mathbf z_y)$, from which the likelihood (or \emph{reconstruction}) term of Equation~\ref{equation:elbo} is computed. The bottom right of the figure finally shows the domain classification head, $\text{NN}_{\bsb \omega_d}$, which takes the domain encoding $\bsb z_d$ of $\bsb x$ as input, and outputs the parameters of the Categorical $q_{\bsb \omega_d}(\mathrm d \vert \mathbf z_d)$, used to compute the domain classification objective $\mathcal L_d$.

Regarding computational cost, assuming that the encoding dimension $M'$ and number of source domains $N_\text{dom}$ are negligible compared to the input dimension $L \cdot M$ (i.e., the network is dominated by its encoder-decoder architecture), \divad\ has the same asymptotic training time as a regular VAE. In practice, \divad\ requires more training resources than VAE, as it uses (i) 2 encoder networks, resulting in twice the encodings and encoder gradients to compute, and (ii) 2 encodings as input to the decoder, leading to more parameters for the first decoder layer.
During inference, \divad\ incurs about half the cost of a VAE, since its anomaly scoring involves only a forward pass through a single encoder, compared to a complete input reconstruction for the VAE. Finally, using \divadgm\ over \divadg\ incurs modest increase in both training and inference costs, with (limited) $K \cdot 2M'$ prior parameters to learn ($2M'$ for each GM component), and $K$ components, instead of 1, to consider when evaluating likelihoods with respect to the prior.

\begin{figure*}
    \begin{center}
	\includegraphics[width=2.0\columnwidth]{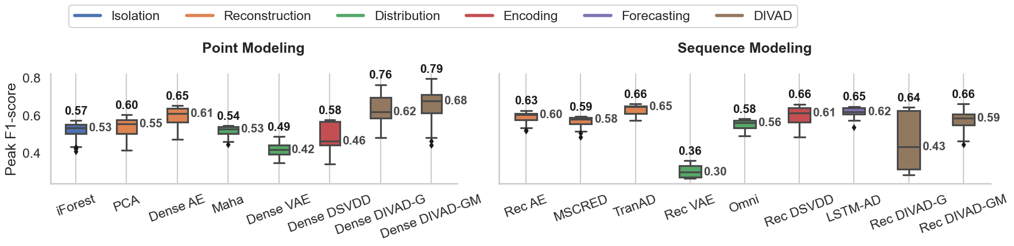}
    \vspace{-0.1in}
    \caption{\small Box plots of peak F1-scores achieved by the existing and \divad\ methods, separated by modeling strategy (point vs. sequence) and colored by method category (from Schmidl et al.~\cite{tsad-eval} plus \divad).}
    \vspace{-0.1in}
    \label{fig:boxplots}
    \Description[Box plots of peak F1-scores achieved by the existing and DIVAD methods, separated by modeling strategy (point vs. sequence) and colored by method category.]{Box plots of peak F1-scores achieved by the existing and DIVAD methods, separated by modeling strategy (point vs. sequence) and colored by method category.}
    \end{center}
\end{figure*}

\section{Experiments}
\label{sec:experiments}
In this section, we evaluate the anomaly detection performance of \divad\ against existing AD methods using both the Exathlon benchmark~\cite{exathlon} and Application Server Dataset (ASD)~\cite{interfusion}.

\subsection{Experimental Setup}
\label{subsec:setup}

Our experimental setup involves  different steps of the Exathlon pipeline~\cite{exathlon}. %
In data preprocessing, we excluded applications 7 and 8, for which there are no disturbed and normal traces, respectively.%
In feature engineering, we dropped the features constant throughout the whole dataset and took the average of   Spark executor features to reduce dimensionality. 
These steps result in  $M=237$ features to use by the AD methods. 

\underline{Data Partitioning.}
To build a single AD model for all Spark applications, we ensure that the 8 Spark applications are represented in both the training and test sets. The training set includes normal traces and some disturbed traces to increase the variety in application \emph{settings} and \emph{input rates}. After data partitioning, our test sequences contain     
15 Bursty Input (\textbf{T1}) anomalies; 
5 Bursty Input Until Crash (\textbf{T2}) anomalies; 
6 Stalled Input (\textbf{T3}) anomalies;
7 CPU Contention (\textbf{T4}) anomalies;
5 Driver Failure (\textbf{T5}) anomalies;
5 Executor Failure (\textbf{T6}) anomalies.

\underline{Training and Inference.}
All deep learning methods use the same random 20\% of training data as validation. By default, they are trained for $300$ epochs, using a Stochastic Gradient Descent (SGD) strategy, mini-batches of size $B$, the AdamW optimizer~\cite{adamw}, a weight decay coefficient of $0.01$, early stopping and checkpointing on the validation loss with a patience of $100$ epochs.

The hyperparameters are treated by following recommended practices~\cite{outlier-book-ch1,sintel}. 
For \emph{architecture parameters}, we start with the default architecture setting of each AD method as suggested in its original paper, and vary the number of hidden units, number of layers or latent dimension by a factor of 2-4, resulting in $n_1$ architectures per method (we generate more model variants for shallow methods, leading to larger values of $n_1$).
Then, the \emph{learning rate} is tuned for each architecture, considering $\eta \in \{1\mathrm{e}{-5}, 3\mathrm{e}{-5}, 1\mathrm{e}{-4}, 3\mathrm{e}{-4}\}$, 
and selecting the value that yields the lowest validation loss (i.e., the best \emph{modeling} performance~\cite{sintel}). The \emph{batch size} is method-dependent and set to the value used in the method's original paper (or if absent, to 32 by default). This entails $n_1$ trained models per method, each with its ``best'' learning rate and recommended batch size.
At inference time (running each model on the test sequences), we derive the record scoring function $g$ using a grid of $12$ \emph{anomaly score smoothing factors} $\smoothing$, leading to  $12n_1$ runs per AD method. We further filter out the architectures whose runs give overall poor performance, resulting in $12n_2$ runs, with $n_2 \le n_1$, per AD method.

We evaluate AD methods based on their \emph{point-based} AD performance, using the \textbf{peak F1-score} metric of the Exathlon benchmark (i.e., the ``best"-possible F1-score on the Precision-Recall curve). When computing F1-scores, we average Recall values across different event types. 
We finally summarize the performance of each AD method, in terms of peak F1-score, using a box-plot over its $12n_2$ model runs. 
Additional details are available in\techreport{~Appendix~\ref{appendix:setup-eval}}{~\cite{tech-report}}.

\underline{Representative AD Methods.}
Our analysis compares DIVAD to 13 unsupervised AD methods, either performing a \emph{point modeling} (i.e., window length $L=1$) or \emph{sequence modeling} (window length $L>1$) of the data. The unsupervised AD methods are further grouped based on their anomaly scoring strategy, as per the taxonomy of Schmidl et al.~\cite{tsad-eval} discussed in Section~\ref{sec:related}.

We include the following \textbf{point modeling} AD methods in our study (details about these methods and hyperparameters considered are given in\techreport{~Appendix~\ref{appendix:point-methods}}{~\cite{tech-report}}):
{\bf (1)}~Isolation forest~\cite{iforest} (\textbf{iForest}), as the most popular isolation tree method;
{\bf (2-3)}~Principal Component Analysis (\textbf{PCA}) \cite{pca} and Dense Autoencoder (\textbf{Dense AE}) \cite{replicator, ae}, as representative and popular shallow and deep reconstruction methods, respectively;
{\bf (4)}~Dense Deep SVDD~\cite{deep-svdd} (\textbf{Dense DSVDD}), as a recent and popular encoding method; %
{\bf (5-6)}~Mahalanobis~\cite{pca, outlier-book-ch3} (\textbf{Maha}) and Dense Variational Autoencoder (\textbf{Dense VAE})~\cite{baseline-vae}, as representative shallow and deep distribution methods, respectively.

We include the following \textbf{sequence modeling} methods (with further details given   in\techreport{~Appendix~\ref{appendix:sequence-methods}}{~\cite{tech-report}}):
{\bf (7-9)}~Recurrent Autoencoder~\cite{replicator, ae} (\textbf{Rec AE}), \textbf{MSCRED}~\cite{mscred} and \textbf{TranAD}~\cite{tranad}, as the sequence modeling version of Dense AE and more recent reconstruction methods, respectively; 
{\bf (10)}~\textbf{LSTM-AD}~\cite{lstm-ad}, as the most popular forecasting method; %
{\bf (11)}~Recurrent Deep SVDD~\cite{deep-svdd} (\textbf{Rec DSVDD}) as the sequence modeling version of the Dense DSVDD encoding method;
{\bf (12-13)}~Recurrent VAE (\textbf{Rec VAE}) and OmniAnomaly~\cite{omni-anomaly} (\textbf{Omni}) as the sequence modeling version of Dense VAE and a more recent distribution method, respectively.
We also tried to include the more recent encoding method DCDetector~\cite{dcdetector}, but did not retain it due to its poor performance on our dataset.

\underline{\divad\ Variants.}
This study considers 4 \divad\ variants, either performing a point modeling or sequence modeling of the data. 
The point modeling variants, referred to as \textbf{Dense \divadg} and \textbf{Dense \divadgm}, use a fully-connected architecture for the encoders $\text{NN}_{\bsb \phi_d}$,  $\text{NN}_{\bsb \phi_y}$ and decoder $\text{NN}_{\bsb \theta_{yd}}$. The sequence modeling variants, referred to as \textbf{Rec \divadg} and \textbf{Rec \divadgm}, use recurrent architectures instead (more details are in\techreport{~Appendix~\ref{appendix:divad}}{~\cite{tech-report}}). We employ the same hyperparameter selection strategy for \divad\ variants as the other DL methods, considering KL divergence weights $\beta \in \{1, 5\}$ (i.e., a regular VAE and $\beta$-VAE~\cite{beta-vae} framework with increased latent space regularization), and a domain classification weight $\alpha_d = 100,000$ set based on the scale we observed for the losses $\mathcal L_{\text{ELBO}}$ and $\mathcal L_d$ in initial experiments. We study the sensitivity of \divad\ to those two hyperparameters in Section~\ref{subsec:sensitivity-beta}.

\vspace{-0.01in}
\subsection{Results and Analyses using Exathlon}
\label{subsec:ad-results}

For the Exathlon benchmark (detailed in \S\ref{sec:ad-analysis}), 
Figure~\ref{fig:boxplots} shows the box plots of the peak F1-scores achieved by the \divad\ variants and 13 AD methods across their hyperparameter values. It separates point from sequence modeling methods into two subplots with a shared y-axis, with boxes colored based on the method category. 

\vspace{-0.05in}
\subsubsection{Existing AD Methods}
\label{subsec:existing-ad}

Our main observations about existing AD methods are the following:
(1) The best performance achieved is the maximum peak F1-score of $0.66$ by TranAD, which is not highly accurate. 
(2) Across different categories, \textit{reconstruction methods performed the best} (with a maximum peak F1-score of $0.66$ by TranAD) \textit{while distribution methods performed the worst} on average (with a maximum peak F1-score of $0.58$ by OmniAnomaly). 
These results are also consistent with the study of~\cite{tranad}, which reported that TranAD outperformed OmniAnomaly and MSCRED. 
(3) The use of deep learning was beneficial among reconstruction methods, while it tended to degrade performance for distribution methods---our subsequent domain shift analysis will explain this behavior.  

\vspace{-0.05in}
\subsubsection{Analysis of Domain Shift}
\label{subsec:shift-analysis}

\begin{figure}
\vspace{-0.1in}
    \centering
    \begin{subfigure}[b]{0.42\textwidth}
        \centering
        \includegraphics[width=\textwidth]{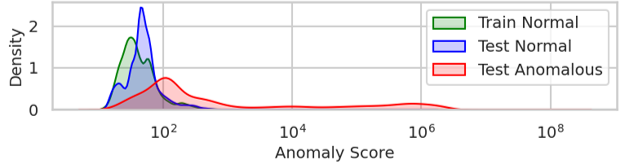}
        \vspace{-0.2in}
        \caption{PCA (shallow reconstruction method).}
        \label{fig:pca-scores}
    \end{subfigure}
    \hfill
    \begin{subfigure}[b]{0.42\textwidth}
        \centering
        \includegraphics[width=\textwidth]{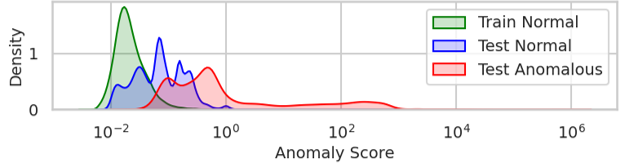}
        \vspace{-0.2in}
        \caption{Dense AE (deep reconstruction method).}
        \label{fig:dense-ae-scores}
    \end{subfigure}
    \begin{subfigure}[b]{0.42\textwidth}
        \centering
        \includegraphics[width=\textwidth]{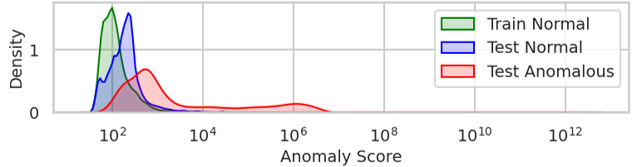}
        \vspace{-0.2in}
        \caption{Maha (shallow distribution method).}
        \label{fig:maha-scores}
    \end{subfigure}
    \hfill
    \begin{subfigure}[b]{0.42\textwidth}
        \centering
        \includegraphics[width=\textwidth]{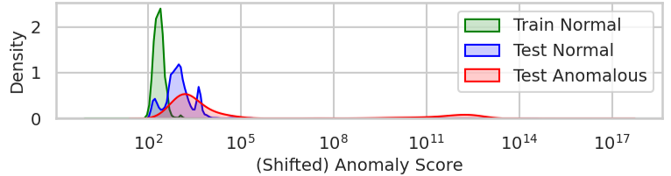}
        \vspace{-0.2in}
        \caption{Dense VAE (deep distribution method).}
        \label{fig:dense-vae-scores}
    \end{subfigure}
    \vspace{-0.1in}
    \caption{\small Kernel Density Estimate (KDE) plots of the anomaly scores assigned by reconstruction and distribution methods to training normal, test normal and test anomalous records.}
    \vspace{-0.1in}
    \Description[Kernel Density Estimate (KDE) plots of the anomaly scores assigned by reconstruction and distribution methods to training normal, test normal and test anomalous records.]{Kernel Density Estimate (KDE) plots of the anomaly scores assigned by reconstruction and distribution methods to training normal, test normal and test anomalous records.}
    \label{fig:reco-dist-scores}
\end{figure}

Figure~\ref{fig:reco-dist-scores} illustrates the impact of domain shift on AD methods by showing the Kernel Density Estimate (KDE) plots of the anomaly scores they assigned to the training normal, test normal, and test anomalous records. On these plots, the separation between the anomaly scores assigned to the test normal and test anomalous records (i.e., between the {\color{NavyBlue} \textbf{blue}} and {\color{Red} \textbf{red}} KDEs) directly relates to the AD performance of a method. As illustrated for the point modeling reconstruction and distribution methods, all the methods have the test normal scores and test anomalous scores overlapping, hence the limited detection accuracy.

\begin{figure}[t]
    \begin{center}
    \includegraphics[width=0.88\columnwidth]{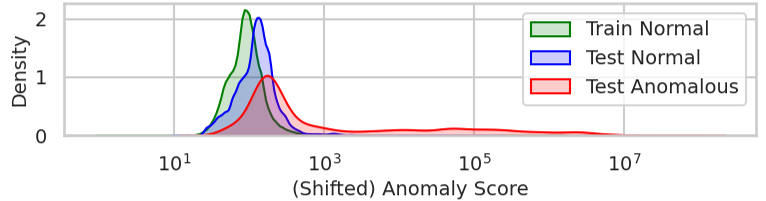}
    \vspace{-0.1in}
    \caption{\small KDE plots of the anomaly scores assigned by Omni to training normal, test normal, and test anomalous records.}
    \label{fig:omni-scores}
    \Description[KDE plots of the anomaly scores assigned by Omni to training normal, test normal, and test anomalous records.]{KDE plots of the anomaly scores assigned by Omni to training normal, test normal, and test anomalous records.}
    \end{center}
    \vspace{-0.2in}
\end{figure}

Furthermore, the overlap between the scores of training normal and test normal records (i.e., {\color{Green} \textbf{green}} and {\color{NavyBlue} \textbf{blue}} KDEs) reflects its ``robustness" to the domain shift from training to test data. 
(1) Comparing Figures~\ref{fig:maha-scores} to~\ref{fig:pca-scores} and~\ref{fig:dense-vae-scores} to~\ref{fig:dense-ae-scores}, respectively, we can see that distribution methods, by modeling the training distribution more \emph{explicitly}, tended to produce more similar anomaly scores across the training normal records (i.e., tighter {\color{Green} \textbf{green}} KDEs). However, this tighter modeling of the training distribution also made these methods more sensitive to domain shift, deeming test normal and test anomalous records ``similarly anomalous'' (i.e., high {\color{NavyBlue} \textbf{blue}} and {\color{Red} \textbf{red}} KDEs overlap), which hindered their performance. 
(2) Comparing Figures~\ref{fig:dense-ae-scores} to~\ref{fig:pca-scores} and~\ref{fig:dense-vae-scores} to~\ref{fig:maha-scores}, we see that deep methods achieved a better separation between the training normal and test anomalous records, by modeling the training data at a finer level than shallow methods. At the same time, they suffer from a larger separation between the training normal and test normal records, indicating their sensitivity to domain shifts. 
In both cases,  \textbf{the more a method precisely and explicitly models the training data, the more vulnerable it is to the domain shift challenge}. 
Figure~\ref{fig:omni-scores} shows a similar plot for OmniAnomaly. We can see that its anomaly scores assigned to training normal records and test anomalies overlapped significantly, indicating a shortcoming in normal data modeling, despite this method being distribution-based and converging properly. Since Omni corresponds to a more advanced extension of VAE (with a non-Gaussian temporal modeling of latent variables), nontrivial extensions could be required to exploit its modeling potential and (at least) match the behavior of Dense VAE.%

\subsubsection{\divad\ vs. Existing AD Methods}
We now examine \divad's performance. First, Figure~\ref{fig:boxplots} shows that \textbf{Dense \divadgm\ and Dense \divadg\ significantly outperform the SOTA method TranAD in maximum performance}, achieving 20\% and 15\% improvements in maximum peak F1-scores ($0.79$ and $0.76$ over $0.66$), respectively. Between the variants, using a learned Gaussian Mixture prior (\divadgm) instead of a fixed Gaussian prior (\divadg) is beneficial in improving both the maximum and median peak F1-scores for the point and sequence modeling variants.

\begin{figure}[t]
    \begin{center}
    \includegraphics[width=0.95\columnwidth]{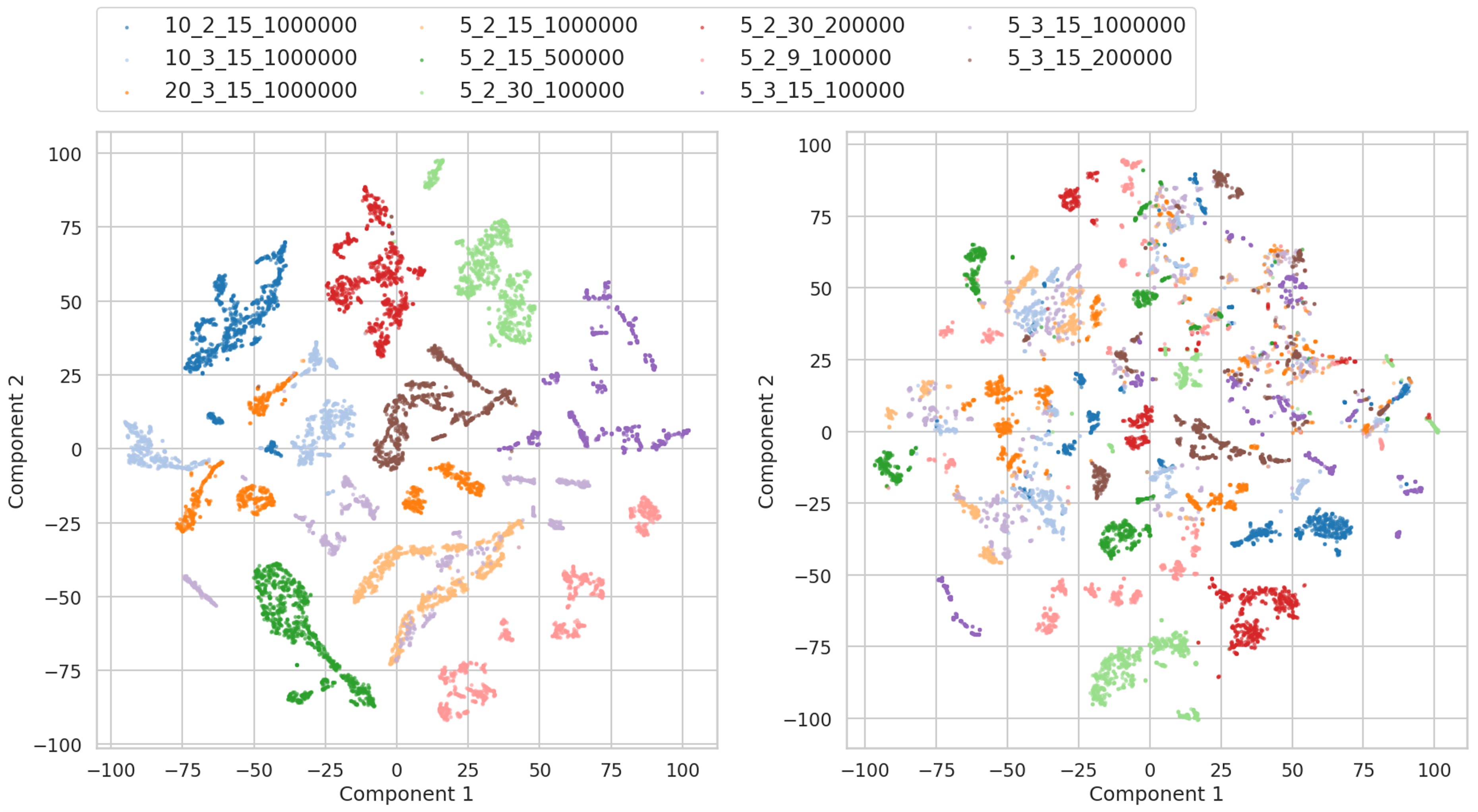}
    \vspace{-0.1in}
    \caption{\small t-SNE scatter plots of Dense \divadgm's domain-specific (left) and domain-invariant (right) encodings of test normal records, undersampled to 10,000 records, balanced and colored by domain.}
    \vspace{-0.2in}
    \label{fig:dense-divad-gm-tsnes}
    \Description[t-SNE scatter plots of Dense DIVAD-GM's domain-specific (left) and domain-invariant (right) encodings of test normal records, undersampled to 10,000 data records, balanced and colored by domain.]{t-SNE scatter plots of Dense DIVAD-GM's domain-specific (left) and domain-invariant (right) encodings of test normal records, undersampled to 10,000 data records, balanced and colored by domain.}
    \end{center}
\end{figure}

Second, \textbf{the higher performance of Dense \divadgm\ can be directly attributed to its accurate domain generalization}. 
To illustrate this, Figure~\ref{fig:dense-divad-gm-tsnes} shows t-SNE scatter plots of the domain-specific and domain-invariant encodings it produced for test normal records (sampled from $q_{\bsb \phi_d}(\mathbf z_d \vert \mathbf x)$ and $q_{\bsb \phi_y}(\mathbf z_y \vert \mathbf x)$, respectively), undersampled to 10,000 data records, balanced and colored by domain. We can see that the mapping learned by Dense \divadgm\ from the \emph{input} to its \emph{domain-specific} space produced the distinct domain clusters expected, while the mapping learned from the \emph{input} to its \emph{domain-invariant} space produced more scattered encodings.

\begin{figure}[t]
    \begin{center}
    \vspace{-0.1in}
    \includegraphics[width=0.88\columnwidth]{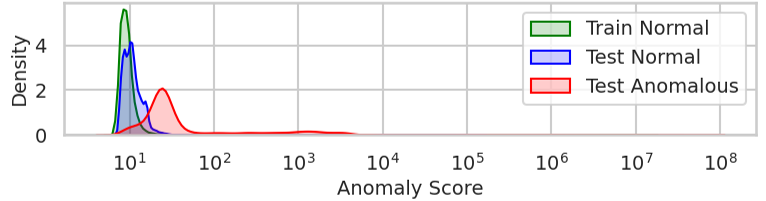}
    \vspace{-0.15in}
    \caption{\small KDE plots of the anomaly scores assigned by Dense \divadgm\ to training normal, test normal, and test anomalous records.}
    \label{fig:dense-divad-gm-scores}
    \Description[KDE plots of the anomaly scores assigned by Dense DIVAD-GM to training normal, test normal and test anomalous records.]{KDE plots of the anomaly scores assigned by Dense DIVAD-GM to training normal, test normal and test anomalous records.}
    \end{center}
    \vspace{-0.1in}
\end{figure}

Furthermore, Figure~\ref{fig:dense-divad-gm-scores} shows the KDE plots of the anomaly scores assigned by the best-performing Dense \divadgm\ to the training normal, test normal and test anomalous records. We see that the explicit modeling of the training data distribution by Dense \divadgm~led to a similar benefit as Dense VAE (see Figure~\ref{fig:dense-vae-scores}), with a low variance in the anomaly scores assigned to the training normal records. Contrary to Dense VAE, Dense \divadgm\ performed this precise density estimation in a \emph{domain-invariant} space (where distribution shifts were drastically reduced), which made it generalize to \emph{test} normal records as well (i.e., better \emph{aligned} and similarly narrow {\color{Green} \textbf{green}} and {\color{NavyBlue} \textbf{blue}} KDEs). As such, Dense \divadgm~could generally view test anomalies as ``more abnormal'' than test normal records, which led to the better performance.

\begin{figure}[t]
	\begin{center}
		\vspace{-0.1in}
		\includegraphics[width=0.95\columnwidth]{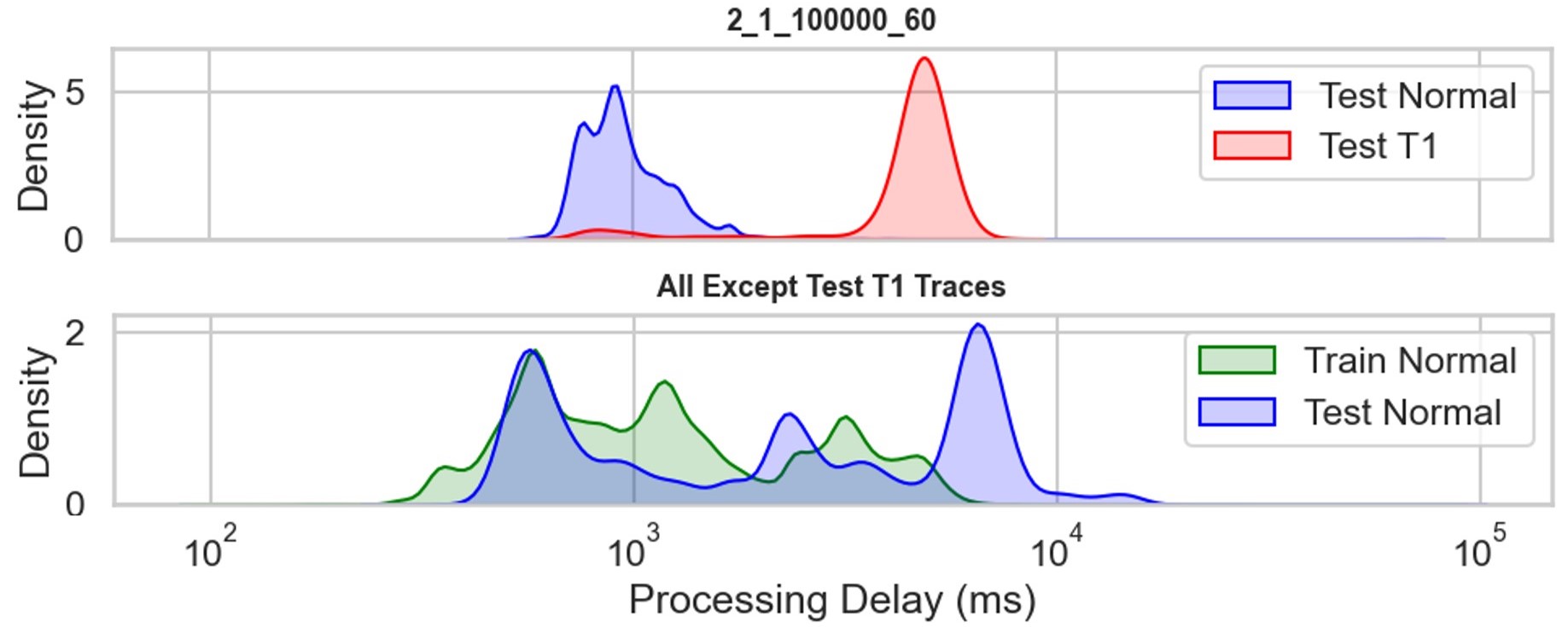}
		\vspace{-0.1in}
		\caption{\small KDE plots of the last completed batch processing delay for training normal data, test normal data, and test anomalous data in a Bursty Input (T1) trace (top) and non-T1 traces (bottom).}
		\vspace{-0.1in}
		\label{fig:t1-contextual}
		\Description[KDE plots of the last completed batch processing delay for training normal data, test normal data and test anomalous data in Bursty Input (T1) traces.]{KDE plots of the last completed batch processing delay for training normal data, test normal data and test anomalous data in Bursty Input (T1) traces.}
	\end{center}
\end{figure}

\subsubsection{\divad\ Variants}

Another observation we can make from Figure~\ref{fig:boxplots} is that \textit{point modeling \divad\ variants could outperform TranAD for our dataset and experimental setup, while sequence modeling variants could not}. Point modeling variants being sufficient here can be explained by the dataset's event types being \textit{mostly reflected as \emph{contextual} anomalies} given our features (i.e., data records that are anomalous in a given context/\emph{domain}, but normal in some others). Figure~\ref{fig:t1-contextual} illustrates this by showing KDE plots of the \textit{last completed batch processing delay} feature over normal data and T1 (Bursty Input) events.  The top plot shows the distributions for a given test T1 trace, while the bottom plot shows them for the remaining data, using the same x-axis in log scale. 
We can see that, although T1 events induce higher processing delays than normal \emph{within the context of a trace}, these ``higher" values actually appear normal with respect to the training and test normal data globally, in particular, in some other \emph{contexts/domains}. When viewed in the domain-invariant spaces of our \divad\ methods, such contextual anomalies could typically be turned into  \emph{point} anomalies (data records deviating from the rest of the data, no matter the context). Referring back to the central assumption of \divad, considering \emph{feature combinations} at \emph{single time steps} at a time was here sufficient for the point modeling methods to learn domain-invariant patterns, given that most anomalies in Exathlon are of the contextual type.

The lower performance observed for sequence modeling \divad\ variants could be explained by the \textit{heightened challenge of learning domain-invariant patterns in the sequential setting}. While leveraging sequential information can be useful in theory, identifying domain-invariant \emph{shapes} within and across $M=237$ time series constitutes a harder task than relying on simple feature combinations at given time steps for our dataset and setup. This can be verified using the anomaly score distributions, with a higher overlap between the test normal and abnormal records explaining the lower performance.

\subsubsection{Sensitivity to Anomaly Scoring Strategy}

\begin{figure}
    \begin{center}
    \includegraphics[width=0.99\columnwidth]{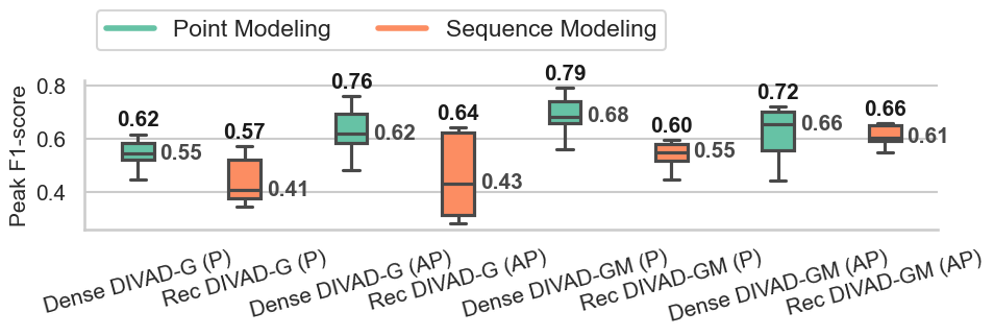}
    \vspace{-0.1in}
    \caption{\small Box plots of peak F1-scores achieved by each DIVAD variant and anomaly scoring strategy (class encoding prior (P) vs. aggregated posterior (AP)), colored by modeling strategy.}
    \vspace{-0.1in}
    \label{fig:dg-scoring-boxplots}
    \Description[Box plots of peak F1-scores achieved by each DIVAD variant and anomaly scoring strategy (class encoding prior vs. aggregated posterior), colored by modeling strategy.]{Box plots of peak F1-scores achieved by each DIVAD variant and anomaly scoring strategy (class encoding prior vs. aggregated posterior), colored by modeling strategy.}
    \end{center}
\end{figure}

Figure~\ref{fig:dg-scoring-boxplots} presents a sensitivity analysis of the anomaly scoring strategy used by our DIVAD methods. It shows the box plots of peak F1-scores achieved by each DIVAD variant and anomaly scoring strategy, with ``(P)" indicating the scoring is based on the class encoding \emph{prior} (fixed Gaussian for \divadg, learned Gaussian Mixture for \divadgm), and ``(AP)" indicating the scoring is based on the class encoding \emph{aggregated posterior} (estimated as a Gaussian for \divadg, and as a Gaussian Mixture with $K$ components for \divadgm). As we can see from this figure, sequence modeling \divad\ methods again performed worse than the point modeling variants in both median and maximum peak F1-scores no matter the scoring strategy used. Like expected, \textbf{deriving the anomaly scores from an aggregated posterior estimate instead of the prior was significantly beneficial for both \divadg\ methods}, which, by relying on a fixed Gaussian prior, are particularly subject to the issue of ``holes in the aggregated posterior" discussed in Section~\ref{subsec:anomaly-scoring}. By relying on a \emph{more expressive} and \emph{learned} class encoding prior, \textbf{\divadgm\ was less sensitive to the type of scoring strategy used}, with the scoring based on the prior performing better in point modeling, and the one based on the aggregated posterior performing better in sequence modeling. This observation is consistent with our expectations, and motivated our choice of including both scoring strategies into the hyperparameters grid of \divadgm\ in our study. 

\subsubsection{Sensitivity to Hyperparameters}
\label{subsec:sensitivity-beta}

\begin{figure}[t]
    \begin{center}
    \includegraphics[width=1.0\columnwidth]{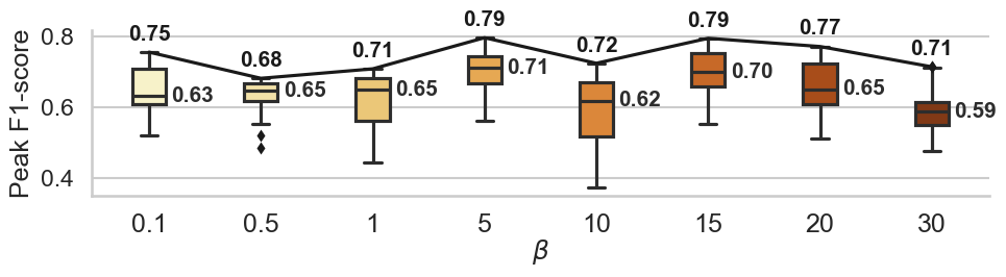}
    \vspace{-0.3in}
    \caption{\small Box plots of peak F1-scores achieved by Dense DIVAD-GM for different KL divergence weights $\beta$.}
    \label{fig:sensitivity-beta}
    \Description[Box plots of peak F1-scores achieved by Dense DIVAD-GM for different KL divergence weights $\beta$.]{Box plots of peak F1-scores achieved by Dense DIVAD-GM for different KL divergence weights $\beta$.}
    \end{center}
    \vspace{-0.1in}
\end{figure}

Figure~\ref{fig:sensitivity-beta} shows the
box plots of peak F1-scores achieved by Dense \divadgm\ across different KL divergence weights $\beta$. From this figure, we can see that finding an optimal $\beta$ value improves both the maximum and median performance significantly (by up to 16\% and 20\%, respectively, from the worst value). The benefit of Dense \divadgm\ over other AD methods, however, remains robust across all $\beta$ values tested (recall that the best peak F1-score of other AD methods is $0.66$).

\begin{figure}[t]  %
    \begin{center}
    \includegraphics[width=1.0\columnwidth]{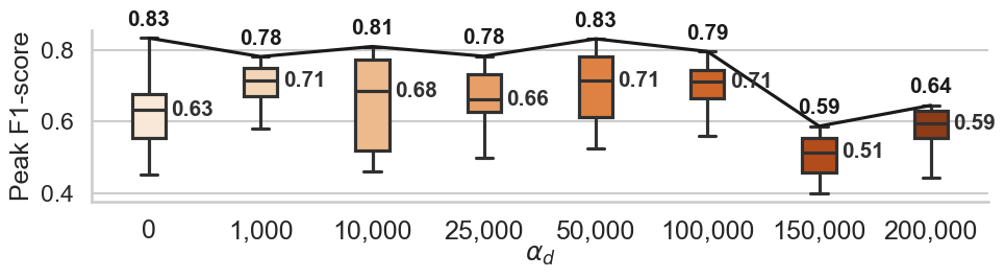}
    \vspace{-0.3in}
    \caption{\small Box plots of peak F1-scores achieved by Dense DIVAD-GM for different domain classification weights $\alpha_d$.}
    \vspace{-0.2in}
    \label{fig:sensitivity-alpha}
    \Description[Box plots of peak F1-scores achieved by Dense DIVAD-GM for different domain classification weights $\alpha_d$.]{Box plots of peak F1-scores achieved by Dense DIVAD-GM for different domain classification weights $\alpha_d$.}
    \end{center}
\end{figure}

Figure~\ref{fig:sensitivity-alpha} shows the peak F1-scores achieved by Dense \divadgm\ across different domain classification weights $\alpha_d$. We see that its maximum peak F1-score is robust across low and medium $\alpha_d$ values, with some even yielding better results than the value of $100,000$ selected for our study. This figure also shows that obtaining the best performance is possible even without domain classification head $\text{NN}_{\bsb \omega_d}$ (i.e., setting $\alpha_d = 0$). However, enforcing domain information via $\text{NN}_{\bsb \omega_d}$ helps reduce Dense \divadgm's sensitivity to other hyperparameters, enabling it to outperform existing methods more \emph{consistently}, with significantly higher median and upper quartile peak F1-scores for suitable $\alpha_d$ values than for $\alpha_d = 0$.

\subsubsection{Training and Inference Times}
\label{subsec:times}
Table~\ref{tab:times} shows the average time of training and inference steps (one step per mini-batch of size $B=32$) for the VAE and \divad\ variants on an NVIDIA A100 80GB PCIe, 
with hyperparameters adjusted to make \divad\ and VAE directly comparable (details are in\techreport{~Appendix~\ref{appendix:times}}{~\cite{tech-report}}). This table shows the expected trend for training: \divad's training steps take about twice the time of VAE's, and \divadgm\ takes 16.5\% longer than \divadg\ on average for a given architecture. During inference, Table~\ref{tab:times} confirms that \divad\ takes less than half the time of VAE, with no significant difference between \divadg\ and \divadgm.

\begin{table}
    \centering
    \small
    \bgroup
    \def\arraystretch{1.1}
    \caption{\small Training and inference times for \divad\ and VAE.}   
    \label{tab:times}
    \vspace{-0.1in}
    \begin{tabular}{|c|c|c|}
    \hline
    \textbf{Method} & \textbf{Training Step (ms)} & \textbf{Inference Step (ms)} \\ \hline
    \textbf{Dense VAE~\cite{baseline-vae}} & 3.3 & 19.4  \\
    \textbf{Dense \divadg} & 7.3  & 9.1 \\
    \textbf{Dense \divadgm} & 9.7  & 8.9 \\ \hline
    \textbf{Rec VAE~\cite{baseline-vae}} & 7.2  & 28.1 \\
    \textbf{Rec \divadg} & 11.6  & 12.9 \\
    \textbf{Rec \divadgm} & 13.3  & 12.4 \\ \hline
    \end{tabular}
    \egroup
\end{table}

\begin{figure}[t]
    \begin{center}
    \vspace{-0.1in}
    \includegraphics[width=0.99\columnwidth]{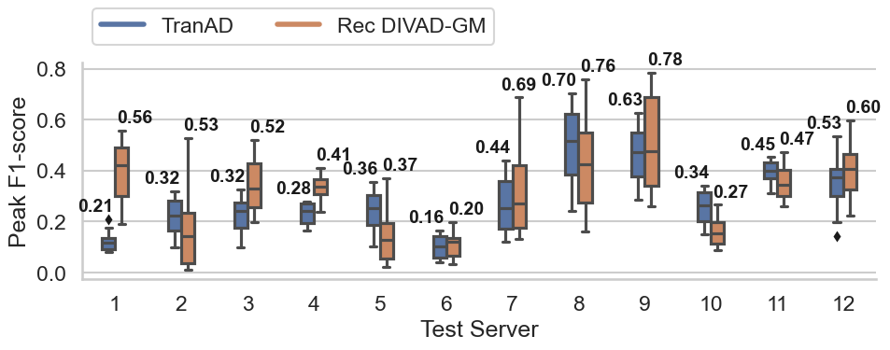}
    \vspace{-0.1in}
    \caption{\small Box plots of peak F1-scores achieved by TranAD and Rec DIVAD-GM for ASD, using each server as a test set.}
    \vspace{-0.1in}
    \label{fig:asd-boxplots}
    \Description[Box plots of peak F1-scores achieved by TranAD and Rec DIVAD-GM for ASD, using each server as a test set.]{Box plots of peak F1-scores achieved by TranAD and Rec \divadgm~for ASD, using each server as a test set.}
    \end{center}
\end{figure}

\begin{figure}[t]
    \centering
    \begin{subfigure}[tb]{0.42\textwidth}
    \centering
    \includegraphics[width=\textwidth]{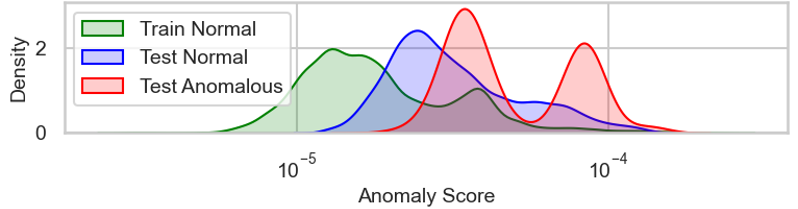}
    \caption{\small TranAD}
    \label{fig:tranad-asd-scores}
    \end{subfigure}
    \begin{subfigure}[tb]{0.42\textwidth}
    \centering
    \includegraphics[width=\textwidth]{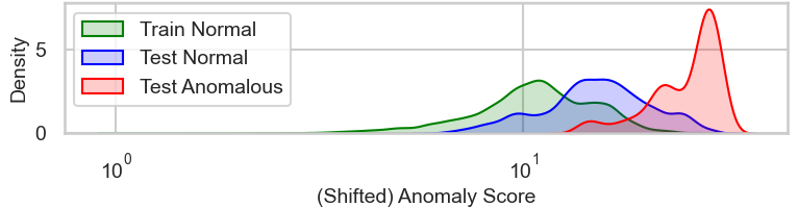}
    \caption{\small Rec \divadgm.}
    \label{fig:rec-divad-asd-scores}
    \end{subfigure}
    \vspace{-0.1in}
    \caption{\small KDE plots of the anomaly scores assigned by TranAD and Rec \divadgm\ to ASD's data records, using  server 1 as test.} %
    \vspace{-0.2in}
    \Description[KDE plots of the anomaly scores assigned by TranAD and Rec \divadgm\ to ASD's data records, using  server 1 as test.]{KDE plots of the anomaly scores assigned by TranAD and Rec \divadgm\ to ASD's data records, using  server 1 as test.}
    \label{fig:asd-scores}
\end{figure}

\subsection{Broader Applicability: ASD Use Case}
\label{sec:asd-dg-experiment}

We now study the broader applicability of our DIVAD framework using the Application Server Dataset (ASD)~\cite{interfusion}. This dataset, collected from a large Internet company, consists of 12  \emph{traces}, each of which recorded the status of a group of services running on a separate \emph{server}, using 19 metrics every five minutes. 
The goal is to detect the labeled anomaly ranges located at the end of the traces. The anomaly ratio is 4.61\%, with minimum, median and maximum anomaly lengths of 3, 18 and 235 data records, respectively. 

In this study, we use ASD to assess the extent to which our \divad\ framework can learn \emph{server-invariant} normal patterns to detect anomalies in a new, \emph{unseen} test server. As such, our experimental setup considers 11 out of the 12 traces as training (without the anomalies) for a single model instance, and the remaining trace as test. 
For these 12 runs, we report the performance of TranAD (the best-performing existing method) and Rec \divadgm. This time, Rec \divadgm\ indeed outperformed Dense \divadgm\ in our experiments, most likely due to (i) a higher presence of \emph{collective} anomalies in ASD (i.e., data records that are anomalous collectively, but not individually), and (ii) the lower number of features $M=19$, making it easier to identify meaningful domain-invariant shapes among them. We consider a window length $L=20$ for both methods, and the same model training and selection strategy as in the previous study. 
More details can be found in\techreport{~Appendix~\ref{appendix:asd}}{~\cite{tech-report}}.

Figure~\ref{fig:asd-boxplots} presents the results of our 12 tests, showing the box plots of peak F1-scores achieved by TranAD and Rec \divadgm\ across their hyperparameter values for each test server. We see that Rec \divadgm\ outperforms TranAD in maximum peak F1-score for 11 out of 12 test servers (i.e., 92\% of the cases), improving the maximum performance by more than 10\% for 8 of them. These results also show that the \emph{median} performance is improved by Rec \divadgm\ for 7 out of the 12 test servers, indicating the sensitivity of \divad\ with respect to hyperparameters, and the necessity of properly tuning them to benefit from a performance gain. 

Figures~\ref{fig:tranad-asd-scores} and~\ref{fig:rec-divad-asd-scores} show the KDE plots of the anomaly scores assigned by the best-performing TranAD and Rec \divadgm\ to training normal, test normal, and test anomalous records when using server 1 as a test set (i.e., the setup for which Rec \divadgm\ improved the performance the most, by 167\%). From Figure~\ref{fig:tranad-asd-scores}, we can see that the low performance of TranAD was primarily to the lower mode of its distribution of anomaly scores assigned to anomalies, which had a significant overlap, and thus were considered ``similarly abnormal", to some test normal data. As shown in Figure~\ref{fig:rec-divad-asd-scores}, Rec \divadgm\ was able to alleviate this issue, producing much less overlap between this lower mode and the rest of test normal data, which explains the performance gain.

\section{Conclusions and Future Directions}
\label{sec:conclusion}

This paper presented a unified framework for benchmarking anomaly detection (AD) methods, and highlighted the problem of \emph{shifts} in normal behavior in practical AIOps scenarios. We then formally formulated the AD problem under domain shift and proposed a new approach, Domain-Invariant VAE for Anomaly Detection (\divad), to learn domain-invariant representations for effective anomaly detection in unseen domains. Evaluation results show that the two main \divad\ variants significantly outperform the best unsupervised AD method using the Exathlon benchmark, with 15-20\% improvements in maximum peak F1-scores, and can be applied to the Application Server Dataset to demonstrate broader applicability.

Our future research directions include a \emph{weakly-supervised} extension of \divad, combining its explicit modeling of normal behavior shifts with a higher robustness to removing anomaly signals enabled by a few training anomalies, and enhancing the model with \emph{explainability}, indicating the reasons behind anomalies, which will be key to widespread adoption in real-world use cases.

\bibliographystyle{ACM-Reference-Format}
\bibliography{refs}

\appendix
\section{Details on the Experimental Setup}
\label{appendix:experimental-setup}

This section details the experimental setup we consider to run and evaluate AD methods on Exathlon's dataset with respect to the problem statement of Section~\ref{sec:problem}. This setup corresponds to the parameters we set for the corresponding steps (and substeps) of the Exathlon pipeline~\cite{exathlon}. 

\subsection{Data Selection}

We never consider the Spark streaming applications 7 and 8, for which there are no disturbed and undisturbed traces, respectively. Our primary goal with this use case is to detect anomalies in the behavior of a running Spark streaming \emph{application}, as opposed to the behavior of the entire four-node \emph{cluster} an application is running on. As such, we always remove from our labels the CPU contention events that had no impact on a recorded application's components (i.e., that occurred on nodes where this application had no running driver or executors). In practice, we label those anomalies the same way as other ``\emph{unknown}" anomalous events, in order not to penalize CPU contention Recall for missing them, nor Precision for detecting records as abnormal in these ranges.

\vspace{-0.05in}
\subsection{Data Preprocessing}
\label{appendix:data-preprocessing}

The metrics collected for Spark streaming traces allocated $140$ columns for each of five ``executor spots" in the data, saved in case one of the two to three active executors of an application failed during its execution. In practice, inactive executor spots in the data took the default value of $-1$ (as a placeholder  for ``null"). This value of $-1$ was however also (and mainly) used to refer to (a potential subset of) ``missing" metrics, not received fast enough for the expected timestamp during data collection. This convention yielded two types of contiguous ``$-1$ ranges" for executor metrics in the data, with some meaning the executor was inactive, and some meaning it was not reachable by data collection (typically, but not only, during anomalies). For every executor \texttt{e}, we distinguish these two cases based on the \texttt{\{e\}\_executor\_runTime\_count} counter metric. Specifically, if a $-1$ range occurs between two non-($-1$) ranges and this counter metric was reset after it, then this $-1$ range corresponds to an ``inactive executor range". Otherwise, it corresponds to a ``missing range". We handle cases of starting and ending $-1$ ranges through a combination of manual inspection and duration rules. With these types of ranges distinguished, missing values were filled by propagating forward preceding valid values (or propagating backward following valid values when no such values existed). Inactive executor ranges were left as $-1$. 

After this preprocessing, all metrics in the data should be either positive or $-1$. This was sometimes not the case for three specific metrics, which could mistakenly take the opposite of their ``true value" due to other related metrics being $-1$. For this reason, we also set all negative metrics different from $-1$ to their opposite values. Finally, to handle duplicate and missing timestamps, we resampled the data from all traces to match their supposed sampling period of one second (using the $\max(\cdot)$ aggregation function). 

\vspace{-0.05in}
\subsection{Data Partitioning}
\label{appendix:data-partitioning}

This paper considers the setup of building a single AD method instance for all the Spark streaming applications in training, as opposed to building a distinct instance per application. We retained this setup to reduce the modeling cost as the number of applications increases. Besides, training a single model for a variety of entities is often considered more effective in practice, allowing this model to share knowledge across entities, and thus increasing data efficiency per entity~\cite{raincoat}. We however do not require AD methods to generalize to \emph{unseen} applications, by making sure the eight applications left after data selection are represented in both training and test. We define our training sequences as most of the undisturbed traces, plus some disturbed traces to increase the variety in application \emph{settings} and \emph{input rates} for the methods to learn from. After data partitioning, our test sequences contain:

\begin{itemize}
    \item \textbf{15} Bursty Input (\textbf{T1}) ranges.
    \item \textbf{5} Bursty Input Until Crash (\textbf{T2}) ranges.
    \item \textbf{6} Stalled Input (\textbf{T3}) ranges.
    \item \textbf{7} CPU Contention (\textbf{T4}) ranges.
    \item \textbf{5} Driver Failure (\textbf{T5}) ranges.
    \item \textbf{5} Executor Failure (\textbf{T6}) ranges.
\end{itemize}

\subsection{Feature Engineering}
\label{appendix:feature-engineering}

All our compared methods consider the same features as input, built from an automated feature engineering step simply consisting in:

\begin{itemize}
    \item Dropping the features collected by \texttt{nmon} (since these features reflect the behavior of the entire four-node cluster, sometimes unrelated to the application run represented in the trace).
    \item Dropping the features that were constant throughout the whole data.
    \item Differencing \emph{cumulative} features (i.e., that were only increasing within a given trace).
    \item Averaging corresponding Spark executor features across ``active executor spots" (non-($-1$) after data preprocessing) into a single block of $140$ features.
\end{itemize}

After this feature engineering, we get $M=237$ features to use by the AD methods, which can be decomposed as follows:

\begin{itemize}
    \item \underline{\textbf{168 Driver Features}}
    \begin{itemize}
        \item 18 ``streaming" features. For example:
        \begin{itemize}
            \item The processing delay and scheduling delay of the last completed batch.
            \item The number of records in the last received batch. 
        \end{itemize}
        \item 5 block manager features. For example:
        \begin{itemize}
            \item The disk space used by the block manager. 
            \item The memory used by the block manager.
        \end{itemize}
        \item 32 JVM features. For example:
        \begin{itemize}
            \item The heap memory usage of the driver.
            \item The survivor space usage of the driver (the survivor space is a memory pool that holds objects having survived a young generation garbage collection, before those objects potentially get promoted to old generation memory).
        \end{itemize}
        \item 19 DAG scheduler features. For example:
        \begin{itemize}
            \item The number of active jobs.
            \item The number of running stages. 
        \end{itemize}
        \item 94 live listener bus features. For example:
        \begin{itemize}
            \item The number of messages received from the DAG scheduler in the last 1, 5 and 15 minutes.
            \item The average processing time of messages received from the DAG scheduler.
        \end{itemize}
    \end{itemize}
    \item \underline{\textbf{69 Executor Features (Averaged Across Active Execs)}}
    \begin{itemize}
        \item 27 ``executor" features. For example:
        \begin{itemize}
            \item The CPU time. 
            \item The number of active tasks.
            \item The number of bytes read and written to HDFS.
        \end{itemize}
        \item 38 JVM features, similar to those of the driver.
        \item 4 netty block transfer features. For example:
        \begin{itemize}
            \item The direct memory used by the shuffle client and server of the netty network application framework (sending and receiving blocks of data).
            \item The heap memory used by the shuffle client and server of the netty network application framework.
        \end{itemize}
    \end{itemize}
\end{itemize}

\subsection{Data Windowing}
\label{appendix:data-windowing}

To compare different AD methods in a unified manner, this paper relies on the framework introduced in Section~\ref{subsec:ad-framework}, only considering methods that are both trained and used on data windows, or \emph{samples}, with anomaly scores derived from a window scoring function $g_W$. We therefore perform a data windowing step, producing sliding windows of length $L=1$ for methods that model individual data records (called \emph{point modeling} methods in the following), and $L=20$ for \emph{sequence modeling} methods.

Once sliding windows have been created from the training sequences, we \emph{balance} them by according to their (application, Spark settings, input rate) triplet for the existing methods. Since there is no reason for AD methods to favor any particular values of those aspects, we indeed ensure every combination that exists in the training data is equally represented. For the \divad\ methods, we balance windows according to their \emph{domain} instead. For both balancing strategies, we make sure this process preserves the data cardinality, by randomly undersampling the over-represented combinations, and randomly oversampling the under-represented ones.

\subsection{AD Inference and Evaluation}
\label{appendix:setup-eval}

When deriving our record scoring functions $g$, we always consider the following grid of anomaly score smoothing factors:

\begin{align*}
\smoothing \in ( & 0, 0.8, 0.9, 0.95, 0.96667, 0.975, \\
            & 0.98, 0.98333, 0.9875, 0.99167, 0.99375, 0.995), 
\end{align*}

\noindent which corresponds to considering approximately the last:

$$
n_{\smoothing} \in (1, 5, 10, 20, 30, 40, 50, 60, 80, 120, 160, 200)
$$

\noindent anomaly scores in the exponentially weighted moving average (with $n_{\smoothing} = 1 / (1 - \smoothing)$).

In Exathlon, online scorers $g$ are evaluated based on their ability to separate normal from anomalous records in anomaly score space, leaving the selection of a suitable threshold to human operators when the solution is deployed in practice~\cite{donut}. A way to do so is by considering every possible detector $f$ that can be derived from $g$ using a fixed anomaly score threshold. That is, given the set of all anomaly scores $g$ assigned in test sequences:

$$
\hat{\mathcal G}_{\text{test}} := \bigcup_{i \in [N_1+1 \isep N_1+N_2]} \left\{\left\{g(\bsb S^{(i)}; L, \smoothing)_t \ , \ t \in [1 \isep T] \right\}\right\},
$$

\noindent consider all detectors $\mathcal F = \{f(\cdot; L, \smoothing, \delta) \ , \ \delta \in  \hat{\mathcal G}_{\text{test}}\}$, where the binary record-wise prediction assigned by a detector $f(\cdot; L, \smoothing, \delta)$ in a sequence $\bsb S$ at time $t$ is defined as:

$$
f(\bsb S; L, \smoothing, \delta)_t := g(\bsb S; L, \smoothing)_t > \delta.
$$

\noindent Plotting the Precision and Recall for every such detector on the test set gives the Precision-Recall (PR) curve. In this paper, we consider the Precision, Recall and F1-score achieved by the detector $f(\cdot; L, \smoothing, \delta^*)$, where $\delta^*$ is the anomaly score threshold that gave the maximum F1-score on the test set (i.e., the ``best" point on the PR-curve). We refer to this latter metric as the \textbf{peak F1-score} achieved by the online scorer $g$, indicating the detection performance this scorer would achieve given the adequate threshold. Like in~\cite{donut}, we indeed favor that AD methods induce a single, high-performing threshold over many ``medium" ones. We consider \emph{point-based} anomaly detection performance, and Recall values averaged across the different event types, deeming them equally important to detect no matter their cardinality in test data.

We benchmark AD methods assuming a purely unsupervised scenario, where labels are not assumed available even for tuning hyperparameters. As such, we report the performance of each method as its full box plot of peak F1-scores achieved across a ``sensible" grid of hyperparameter values, like advised for instance in~\cite{outlier-book-ch1}.

For every AD method, our window-based methodology to derive the record scoring function can induce ``rightfully large" anomaly scores assigned to the $L-1$ records immediately following an anomaly (since the windows of length $L$ used to compute them are partially anomalous). This may introduce some rightful ``lags" in the anomaly predictions, hindering the global performance despite the method behaving properly. We handle this aspect by ignoring the $L-1$ records following each test anomaly in our evaluation, where $L$ is the window length used by the AD method. 

\section{Details on the Existing Methods}
\label{appendix:methods}

This section provides details about the existing methods compared, covering the model training and selection strategy we used for deep learning methods, as well as a short description and the hyperparameter grid considered for each method.

\subsection{Model Training and Selection for Deep Learning Methods}
\label{appendix:dl-training}

All of the deep learning methods considered used the same random 20\% of training data as validation, sampled in a stratified manner with traces as strata, with the labeled training and validation anomalies removed. Unless mentioned otherwise, all deep learning methods were trained with a Stochastic Gradient Descent (SGD) strategy, using mini-batches of size $B$, the AdamW optimizer~\cite{adamw}, and a weight decay coefficient of $0.01$. For all methods and sets of hyperparameters, we considered a grid of learning rate values $\eta \in \{1\mathrm{e}{-5}, 3\mathrm{e}{-5}, 1\mathrm{e}{-4}, 3\mathrm{e}{-4}\}$, and selected the learning rate that yielded the lowest validation loss (i.e., the best \emph{modeling} performance, like in~\cite{sintel}). All methods were trained for $300$ epochs by default, using early stopping and checkpointing on the validation loss with a patience of $100$ epochs.

\subsection{Point Modeling Methods}
\label{appendix:point-methods}

Point modeling methods model individual data records ($L=1$), assumed independent and identically distributed (i.i.d). As such, they only rely on our feature engineering and anomaly score smoothing to capture the sequential aspect of the data. We include the following point modeling methods in our study:

\begin{itemize}
    \item Isolation forest~\cite{iforest} (\textbf{iForest}) as an isolation tree method.
    \item Principal Component Analysis (\textbf{PCA}) \cite{pca} and Dense Autoencoder (\textbf{Dense AE}) \cite{replicator, ae} as reconstruction methods.
    \item Dense Deep SVDD~\cite{deep-svdd} (\textbf{Dense DSVDD}) as an encoding method.
    \item Mahalanobis~\cite{pca, outlier-book-ch3} (\textbf{Maha}) and Dense Variational Autoencoder (\textbf{Dense VAE})~\cite{baseline-vae} as distribution methods.
\end{itemize}

Isolation forest trains an ensemble of trees to isolate the samples in the training data, and defines the anomaly score of a test instance proportionally to the average path length required to reach it using the trees. We report its performance with the following hyperparameters (using the default values of Scikit-Learn 1.0.2~\cite{scikit-learn} for the ones not mentioned):

\begin{itemize}
    \item A number of trees in $\{50, 100, 200, 500, 1000\}$.
    \item A maximum number of samples used by each tree in $\{256, \\ 512, 2048, 8192, 32768\}$.
    \item A maximum number of features used by each tree of $64$.
\end{itemize}

As reconstruction methods, PCA and Dense AE define anomaly scores of test vectors as their mean squared reconstruction errors from a transformed (latent) space. The transformation of PCA is a projection on the linear hyperplane formed by the principal components of the data. We report its performance with the following preprocessing and hyperparameters (using the default values of Scikit-Learn 1.0.2 for the ones not mentioned):

\begin{itemize}
    \item A standardization of the input samples.
    \item A number of principal components (latent dimension) in $\{16, 64, 128, 95\%, 99\%, M\}$, where $95\%$ and $99\%$ correspond to the latent dimension preserving $95\%$ and $99\%$ of the training data variance, respectively, and $M=237$ is our input dimensionality after feature engineering.
\end{itemize}

The transformation of the Autoencoder method is a non-linear mapping to a latent encoding learned by a neural network that was trained to reconstruct input data from it. With Dense AE, we consider a fully-connected architecture for this neural network, and report its performance with the following preprocessing and hyperparameters:

\begin{itemize}
    \item A standardization of the input samples.
    \item A single hidden layer of $200$ units for both the encoder and the decoder.
    \item The Rectified Linear Unit (ReLU) activation function for all the layers except the output, for which we do not use any activation function.
    \item An encoding dimension in $\{16, 64\}$.
    \item A batch size $B=32$.
\end{itemize}

The Mahalanobis and VAE methods define anomaly scores of data samples as their deviation from an estimated data distribution. The Mahalanobis method estimates this distribution as a multivariate Gaussian, and defines the anomaly score of a test vector as its squared Mahalanobis distance from it. As such, it does not require any hyperparameters, and we therefore report its performance using only a standardization of the input samples. Dense VAE estimates the data distribution using a fully-connected variational autoencoder, with the anomaly score of a test point derived by drawing multiple samples from the probabilistic encoder, and averaging the negative log-likelihood of the reconstructions obtained from each of these samples. We report its performance using the following preprocessing and hyperparameters:

\begin{itemize}
    \item A standardization of the input samples.
    \item A single hidden layer of $200$ units for both the encoder and the decoder.
    \item An encoding dimension in $\{16, 64\}$.
    \item The ReLU activation function for all the layers except the encoding and output. To improve numerical stability, we adopt a similar strategy to Xu et al.~\cite{donut}, and derive the standard deviations of encodings and outputs using softplus activations shifted by a small constant $\epsilon$ set to $1\mathrm{e}{-4}$.
    \item A batch size $B=32$.
    \item A number of samples drawn of $256$ to derive the anomaly score of a test example.
\end{itemize}

The Dense DSVDD method trains a fully-connected neural network to map the input data to a latent representation enclosed in a small hypersphere, and then defines anomaly scores of test samples as their squared distance from this hypershere's centroid. We use the implementation of Ruff et al.~\cite{deep-sad} with the ``One-Class Deep SVDD" objective (assuming most of the training data is normal) and their initialization of the encoder weights from a pretrained autoencoder model. We report the performance of Dense DVSDD with the following preprocessing and hyperparameters (using the same values as the original implementation for those not mentioned):

\begin{itemize}
    \item A standardization of the input samples (we also tried the original paper's strategy of normalizing the inputs and using a sigmoid activation function for the output layer, but this did not lead to a better performance). 
    \item A single hidden layer of $200$ units for the encoder and the decoder (with the decoder only being used for pretraining).
    \item An encoding dimension in $\{16, 64\}$.
    \item The Leaky ReLU activation function with a negative slope coefficient $\alpha = 0.01$ for all the layers except the encoding and decoder output (like in the original implementation).
    \item A batch size $B=200$ (like in the original implementation).
    \item A pretraining phase of $150$ epochs, followed by a training phase of $150$ epochs (which makes the same total number of $300$ epochs as the other methods).
    \item The same learning rate and optimization strategy for the pretraining and training phases.
    \item The same grid of learning rate values as the other methods, but dividing the learning rate by $10$ after $50$ epochs (like in the original implementation). Larger learning rate values were also tried due to this scheduling, but did not produce better results.
    \item A weight decay coefficient of $1\mathrm{e}{-6}$ (like in the original implementation).
\end{itemize}

\subsection{Sequence Modeling Methods}
\label{appendix:sequence-methods}

Sequence modeling methods model wider \emph{windows} of data records ($L = 20$ here), which offers them the capacity of explicitly considering the temporal aspect of the data. We include the following sequence modeling methods in our study:

\begin{itemize}
    \item Recurrent Autoencoder~\cite{replicator, ae} (\textbf{Rec AE}), \textbf{MSCRED}~\cite{mscred} and \textbf{TranAD}~\cite{tranad}, as the sequence modeling version of Dense AE and more recent reconstruction methods, respectively. 
    \item \textbf{LSTM-AD}~\cite{lstm-ad}, as the most popular forecasting method. %
    \item Recurrent Deep SVDD~\cite{deep-svdd} (\textbf{Rec DSVDD}) as the sequence modeling version of the Dense DSVDD encoding method. %
    \item Recurrent VAE (\textbf{Rec VAE}) and OmniAnomaly~\cite{omni-anomaly} (\textbf{Omni}) as the sequence modeling version of Dense VAE and a more recent distribution method, respectively.
\end{itemize}

\begin{figure*}[t]
    \begin{center}
    \includegraphics[width=1.9\columnwidth]{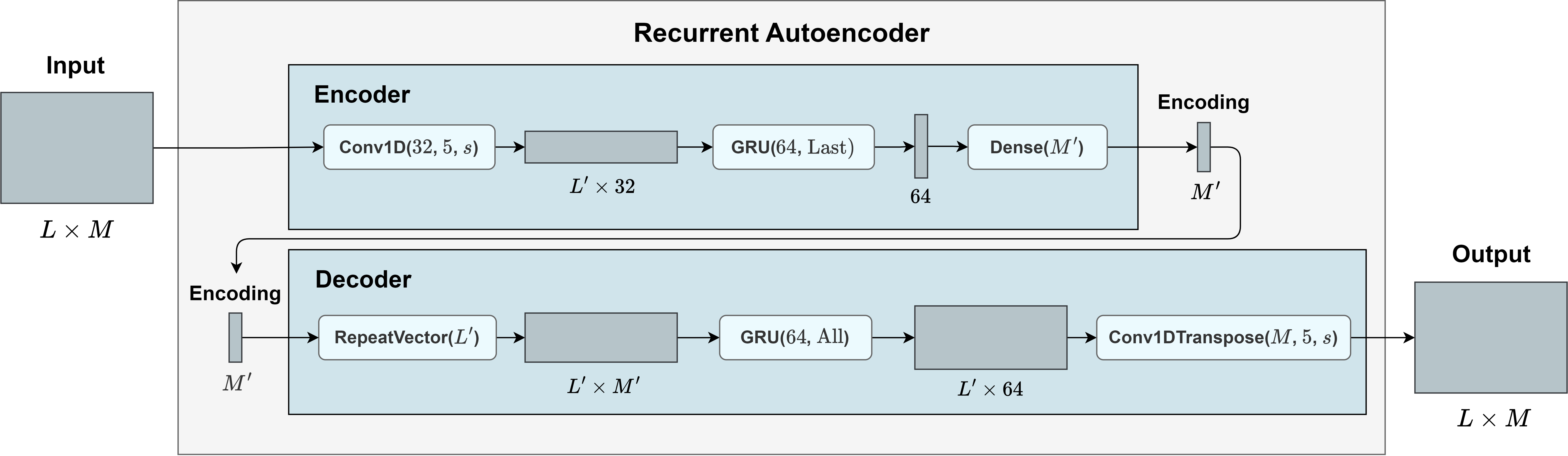}
    \caption{General form of our recurrent autoencoder architectures.}
    \label{fig:conv-rec-ae}
    \Description[General form of our recurrent autoencoder architectures.]{General form of our recurrent autoencoder architectures.}
    \end{center}
\end{figure*}

Rec AE uses the same modeling and scoring strategy as Dense AE, with the fully-connected neural network architecture replaced by a recurrent one. Figure~\ref{fig:conv-rec-ae} illustrates the general form we adopted for our recurrent autoencoders, including 1D convolutional and recurrent layers. In this design, the encoder first consists of an optional stack of 1D convolutional layers, which in this example contains a single layer labeled \texttt{Conv1D}$(32, 5, s)$, to indicate it has 32 filters of size 5 and a stride length hyperparameter $s$. These layers result in a new latent window length $L' \leq L$ for an input window, with one feature map per filter in the last layer. These feature maps get sent to an optional stack of GRU layers, here shown as a single layer labeled \texttt{GRU}$(64, \text{Last})$, to indicate it has 64 units and returns its outputs for the last time step only. These layers are followed by a fully-connected layer that outputs the final encoding. This encoding is provided as input to the decoder, which repeats it $L'$ times to match the window length of the data after the 1D convolutions. This repeated vector goes through a stack of GRU layers typically defined symmetrically to the encoder's, except it now returns its outputs for each of the $L'$ time steps. These outputs finally get passed to a stack of 1D transposed convolutional layers defined symmetrically to the encoder's 1D convolutional layer stack, except for the output layer using $M$ filters to match the input dimensionality. We report the performance of Rec AE using the following preprocessing and hyperparameters:

\begin{itemize}
    \item A standardization of the input samples.
    \item An encoder with a 1D convolutional layer using $32$ filters of size $5$, a stride length of $1$ and the ReLU activation function, followed by a GRU layer of $64$ units using the hyperbolic tangent ($\tanh$) activation function, and a fully-connected layer to output the encoding. 
    \item A decoder defined symmetrically to the encoder as per the design of Figure~\ref{fig:conv-rec-ae}.
    \item An encoding dimension in $\{64, 128\}$, with the ReLU activation function for the encoding layer.
    \item A batch size $B=32$.
\end{itemize}

TranAD uses a transformer-based model with self-conditioning, an adversarial training procedure and model-agnostic meta learning (MAML). It relies on two encoder-decoder networks, with the first encoder considering the current input window, and the second one considering a larger \emph{context} of past data in the window's sequence. The method defines the anomaly score of an input window as the average of its reconstruction errors coming from two decoders and inference phases, with the second phase using the reconstruction error from the first phase as a focus score to detect anomalies at a finer level. Compared to the other methods, TranAD therefore considers training windows augmented with their past sequence data, which prevented us from applying our simple window balancing strategy. We report the performance of TranAD using the implementation of Tuli et al.~\cite{tranad}, only specifying manually the following preprocessing and hyperparameters: 

\begin{itemize}
    \item A normalization of the input samples (like in the original implementation). We also tried the strategy of standardizing the inputs and using no activation function for the output layer (like the other methods), but this did not lead to a better performance.
    \item A number of encoder hidden units in $\{64, 128\}$.
    \item The same grid of learning rate values as the other methods, but multiplying the learning rate by $0.9$ every $5$ epochs (like in the original implementation). Larger learning rate values were also tried due to this scheduling, but did not produce better results.
    \item A batch size $B=128$ (like in the original implementation).
    \item A weight decay coefficient of $1\mathrm{e}{-5}$ (like in the original implementation).
\end{itemize}

MSCRED turns a multivariate time series into multi-scale signature matrices characterizing system status at different time steps, and learns to reconstruct them using convolutional encoder-decoder and attention-based ConvLSTM networks. We report the performance of MSCRED using the implementation from TimeEval~\cite{time-eval}, only specifying manually the following preprocessing and hyperparameters:

\begin{itemize}
    \item A standardization of the input samples.
    \item Two signature matrices of lengths $w = 5, 10$ at each time step, with no gap between consecutive computations.
    \item A number of convolutional encoder layers in $\{2, 3\}$, of the form $((32, 8, 8), (128, 8, 8))$ in case of 2 layers, and $((32, 8, 8),\allowbreak (64, 4, 4), (128, 4, 2))$ in case of 3 layers, with $(f, k, s)$ the number of filters, filter size and stride length of a layer, respectively.
    \item The attention-based ConvLSTM and convolutional decoder networks derived from the convolutional encoder like in the original implementation.
    \item A batch size $B=64$.
    \item A number of epochs of $20$, with an early stopping patience of $10$ epochs.
\end{itemize}

LSTM-AD trains a stacked LSTM network to predict the next $l$ data records from the first $L-l$ of a window. Originally designed for univariate time series, this method produces an $l$-dimensional vector of forecasting errors for each data record in a test sequence, with one component for each position held by this record in forecast windows of length $l$. The method then fits a multivariate Gaussian distribution to the error vectors it produced in a validation set, and defines the anomaly score of a test record as the negative log-likelihood of its error with respect to this distribution. In this work, we adapt LSTM-AD to multivariate data by considering the $l \times M$ matrix of forecasting errors made for a data record at each time step and feature, and averaging these errors across the feature dimension to get the $l$-dimensional vector of the original method. We report the performance of LSTM-AD using the following hyperparameters:

\begin{itemize}
    \item A standardization of the input samples.
    \item A forecast window length $l=10$ (i.e., half of our window length $L=20$).
    \item A version with a single LSTM layer of $128$ units, and a version with two LSTM layers of $128$ units each, all using the $\tanh$ activation function.
    \item A batch size $B=32$.
\end{itemize}

Rec DSVDD uses the same modeling and scoring strategy as Dense DSVDD, with the fully-connected neural network architecture replaced by a recurrent one, adapting the implementation of Ruff et al.~\cite{deep-sad} to match the design of Figure~\ref{fig:conv-rec-ae}. We report the performance of Rec DSVDD with the following preprocessing and hyperparameters (using the same values as the original implementation for those not mentioned):

\begin{itemize}
    \item A standardization of the input samples.
    \item An encoder with a 1D convolutional layer using $32$ filters of size $5$, a stride length of $1$, batch normalization and the Leaky ReLU activation function (with a negative slope coefficient $\alpha = 0.01$), followed by a GRU layer of $64$ units and a fully-connected layer to output the encoding. 
    \item A pretraining decoder defined symmetrically to the encoder as per the design of Figure~\ref{fig:conv-rec-ae}.
    \item An encoding dimension in $\{64, 128\}$, with no activation function for the encoding layer.
    \item A batch size $B=200$ (like in the original implementation).
    \item A pretraining phase of $150$ epochs, followed by a training phase of $150$ epochs.
    \item The same learning rate and optimization strategy for the pretraining and training phases.
    \item The same grid of learning rate values as the other methods, but dividing the learning rate by $10$ after $50$ epochs (like in the original implementation).
    \item A weight decay coefficient of $1\mathrm{e}{-6}$ (like in the original implementation).
\end{itemize}

Rec VAE uses the same modeling and scoring strategy as Dense VAE, with the fully-connected neural network architecture replaced by a recurrent one following the design of Figure~\ref{fig:conv-rec-ae}. We report the performance of Rec VAE using the following preprocessing and hyperparameters:

\begin{itemize}
    \item A standardization of the input samples.
    \item An encoder with a 1D convolutional layer using $32$ filters of size $5$, a stride length of $1$ and the ReLU activation function, followed by a GRU layer of $64$ units using the $\tanh$ activation function, and a fully-connected layer to output the encoding parameters. 
    \item A decoder defined symmetrically to the encoder as per the design of Figure~\ref{fig:conv-rec-ae}.
    \item An encoding dimension in $\{64, 128\}$.
    \item The same strategy as Dense VAE for deriving the encoding and output standard deviations.
    \item A batch size $B=32$.
    \item A number of samples drawn of $256$ to derive the anomaly score of a test example.
\end{itemize}

OmniAnomaly~\cite{omni-anomaly} estimates the distribution of multivariate windows with a stochastic recurrent neural network, explicitly modeling temporal dependencies among variables through a combination of GRU and VAE. It then defines a test window's anomaly score as the negative log-likelihood of its reconstruction. We report the performance of OmniAnomaly using the implementation of Su et al.~\cite{omni-anomaly}, only specifying manually the following preprocessing and hyperparameters: 

\begin{itemize}
    \item A normalization of the input samples (like in the original implementation). We also tried the strategy of standardizing the inputs (like the other methods), but this did not lead to a better performance.
    \item A number of units of $200$ for the fully-connected and GRU layers.
    \item A number of planar normalizing flow layers of $10$.
    \item An encoding dimension in $\{64, 128\}$.
    \item An L2 regularization coefficient of $1\mathrm{e}{-4}$ for all layers (like in the original implementation).
    \item A batch size $B=64$.
    \item The Adam optimizer with a grid $\{3\mathrm{e}{-5}, 1\mathrm{e}{-4}, 3\mathrm{e}{-4}, 1\mathrm{e}{-3}\}$ for the initial learning rate, dividing the learning rate by $2$ every $20$ epochs. 
    \item A gradient norm limit of $10.0$ (like in the original implementation).
    \item A number of epochs of $40$, with an early stopping patience of $10$ validations. To accelerate computations, we run validation $5$ times per epoch, instead of once every $100$ steps in the original implementation.
\end{itemize}

\section{Details on the \divad\ variants}
\label{appendix:divad}

This section provides details about our \divad\ variants, covering our model training and selection strategy, as well as the architecture and hyperparameter grid considered for each variant.

We adopt the same model training and selection strategy as described in Appendix~\ref{appendix:dl-training} for both \divadg~and \divadgm. Like for DIVA~\cite{diva}, we do not share the parameters of our encoder networks $\text{NN}_{\bsb \phi_y}$ and $\text{NN}_{\bsb \phi_d}$, but consider the \emph{multi-encoder} architecture illustrated in Figure~\ref{fig:divad-multi-encoder}. For both variants, we use the same strategy as Dense VAE and Rec VAE for deriving encoding and output standard deviations.

We first consider \emph{point modeling} \divad\ variants ($L=1$), using fully-connected neural network architectures and referred to as \textbf{Dense \divadg} and \textbf{Dense \divadgm}, respectively. We report the performance of Dense \divadg~using the following preprocessing and hyperparameters:

\begin{itemize}
    \item A standardization of the input samples.
    \item A single hidden layer of $200$ units for the encoders $\text{NN}_{\bsb \phi_y}$ and $\text{NN}_{\bsb \phi_d}$ as well as for the decoder $\text{NN}_{\bsb \theta_{yd}}$.
    \item A single fully-connected hidden layer of $64$ units for the conditional domain prior network $\text{NN}_{\bsb \theta_{d}}$.
    \item A ReLU activation function followed by a single fully-con\-nected layer of $N_{\text{dom}} = 22$ units (i.e., one per source domain) for the classification head $\text{NN}_{\bsb \omega_{d}}$. The domain encoding $\bsb z_d$ is indeed sampled unbounded from $q_{\bsb \phi_d}(\mathbf z_d \vert \bsb x)$, so we pass it through a ReLU activation before applying the output layer. The classification head was intentionally kept simple, so as to be able to \emph{easily} classify the domain from $\bsb z_d$.
    \item The ReLU activation function for all the layers except the encoding and output layers.
    \item An encoding dimension in $\{16, 64\}$.
    \item A KL divergence weight $\beta \in \{1, 5\}$.
    \item A domain classification weight $\alpha_d = 100,000$. We set this weight based on the scale we observed for our losses $\mathcal L_{\text{ELBO}}$ and $\mathcal L_d$ during training in some initial experiments. We also tried the CoV-Weighting strategy proposed by Groenendijk et al.~\cite{cov-weighting} to automatically balance these two losses, but this did not lead to a better performance.
    \item A batch size $B=128$.
    \item An anomaly scoring based on the class encoding aggregated posterior estimate $\hat q_{\bsb \phi_y}(\mathbf z_y)$, fitting a multivariate Gaussian distribution to the training class encodings.
\end{itemize}

\noindent We report the performance of Dense \divadgm~using the following preprocessing and hyperparameters:

\begin{itemize}
    \item A standardization of the input samples.
    \item The same architectures as Dense \divadg~for the networks $\text{NN}_{\bsb \phi_y}$, $\text{NN}_{\bsb \phi_d}$, $\text{NN}_{\bsb \theta_{yd}}$, $\text{NN}_{\bsb \theta_{d}}$ and $\text{NN}_{\bsb \omega_{d}}$.
    \item An encoding dimension in $\{16, 32\}$, with $K=8$ Gaussian Mixture components when using an encoding dimension of $16$, and $K=4$ components when using an encoding dimension of $32$.
    \item A KL divergence weight $\beta \in \{1, 5\}$. 
    \item A domain classification weight $\alpha_d = 100,000$. 
    \item A batch size $B=128$.
    \item An anomaly scoring based on (i) the learned class encoding prior $p_{\bsb \lambda}(\mathbf z_y)$, and (ii) the class encoding aggregated posterior estimate $\hat q_{\bsb \phi_y}(\mathbf z_y)$, fitting a Gaussian Mixture distribution with $K$ components to the training class encodings.
\end{itemize}

We also consider \emph{sequence modeling} \divad\ variants ($L=20$ here), with some fully-connected neural network architectures replaced by recurrent ones based on the design of Figure~\ref{fig:conv-rec-ae}, referred to as \textbf{Rec \divadg} and \textbf{Rec \divadgm}, respectively. We report the performance of Rec \divadg~using the following preprocessing and hyperparameters:

\begin{itemize}
    \item A standardization of the input samples.
    \item Each encoder $\text{NN}_{\bsb \phi_y}$ and $\text{NN}_{\bsb \phi_d}$ with a 1D convolutional layer using $64$ filters of size $5$, a stride length of $1$ and the ReLU activation function, followed by a GRU layer of $64$ units using the $\tanh$ activation function, and a fully-connected layer to output the encoding parameters. 
    \item The decoder $\text{NN}_{\bsb \theta_{yd}}$ defined symmetrically to one encoder as per the design of Figure~\ref{fig:conv-rec-ae}.
    \item The same architectures as Dense \divadg~for the conditional domain prior network $\text{NN}_{\bsb \theta_{d}}$ and classification head $\text{NN}_{\bsb \omega_{d}}$.
    \item An encoding dimension of $32$.
    \item A KL divergence weight $\beta \in \{1, 5\}$. 
    \item A domain classification weight $\alpha_d = 100,000$. 
    \item A batch size $B=128$.
    \item The same anomaly scoring as Dense \divadg.
\end{itemize}

\noindent We report the performance of Rec \divadgm~using the following preprocessing and hyperparameters:

\begin{itemize}
    \item A standardization of the input samples.
    \item The same architectures as Rec \divadg~for the networks $\text{NN}_{\bsb \phi_y}$, $\text{NN}_{\bsb \phi_d}$, $\text{NN}_{\bsb \theta_{yd}}$, $\text{NN}_{\bsb \theta_{d}}$ and $\text{NN}_{\bsb \omega_{d}}$.
    \item An encoding dimension of $32$, with $K=8$ Gaussian Mixture components.
    \item A KL divergence weight $\beta \in \{1, 5\}$. 
    \item A domain classification weight $\alpha_d = 100,000$. 
    \item A batch size $B=128$.
    \item The same anomaly scoring as Dense \divadgm.
\end{itemize}

\section{Additional Analyses on Exathlon}
\label{appendix:analyses}

This section provides additional analyses of the results obtained using the Exathlon benchmark.

\begin{figure}[t]
	\begin{center}
		\includegraphics[width=0.82\columnwidth]{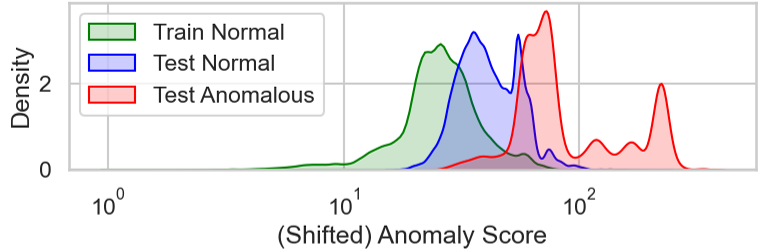}
		\caption{\small KDE plots of the anomaly scores assigned by Rec \divadgm\ to training normal, test normal, and test anomalous records.}
		\label{fig:rec-divad-gm-scores}
		\Description[KDE plots of the anomaly scores assigned by Rec DIVAD-GM to training normal, test normal and test anomalous records.]{KDE plots of the anomaly scores assigned by Rec DIVAD-GM to training normal, test normal and test anomalous records.}
	\end{center}
\end{figure}

\begin{figure}[t]
	\begin{center}
		\vspace{-0.1in}
		\includegraphics[width=0.82\columnwidth]{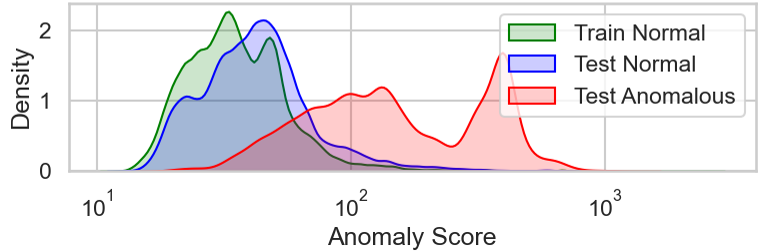}
		\vspace{-0.1in}
		\caption{\small KDE plots of the anomaly scores assigned by Rec \divadg\ to training normal, test normal, and test anomalous records.}
		\label{fig:rec-divad-g-scores}
		\Description[KDE plots of the anomaly scores assigned by Rec DIVAD-G to training normal, test normal and test anomalous records.]{KDE plots of the anomaly scores assigned by Rec DIVAD-G to training normal, test normal and test anomalous records.}
	\end{center}
	\vspace{-0.2in}
\end{figure}

The lower performance observed for sequence modeling \divad\ variants could be explained by the \textit{heightened challenge of learning domain-invariant patterns in the sequential setting}. While leveraging sequential information can be useful in theory, identifying domain-invariant \emph{shapes} within and across $M=237$ time series constitutes a harder task than relying on simple feature combinations at given time steps for our dataset and setup. This is illustrated in Figure~\ref{fig:rec-divad-gm-scores}, showing KDE plots of Rec \divadgm's anomaly scores for training normal, test normal and test anomalous records. From this figure, we can see that the domain generalization performed by Rec \divadgm\ was less effective than for Dense \divadgm\ and Rec \divadg\ (see Figures~\ref{fig:dense-divad-gm-scores} and \ref{fig:rec-divad-g-scores}, respectively), with its anomaly scores \emph{drifting} from the training to the test normal records ({\color{Green} \textbf{green}} vs. {\color{NavyBlue} \textbf{blue}} KDEs). This suboptimal domain generalization led to a higher overlap between the anomaly scores of the test normal and anomalous records ({\color{NavyBlue} \textbf{blue}} vs. {\color{Red} \textbf{red}} KDEs), explaining the lower performance.

\begin{figure}
	\centering
	\begin{subfigure}[b]{1.0\columnwidth}
			\centering
			\includegraphics[width=\columnwidth]{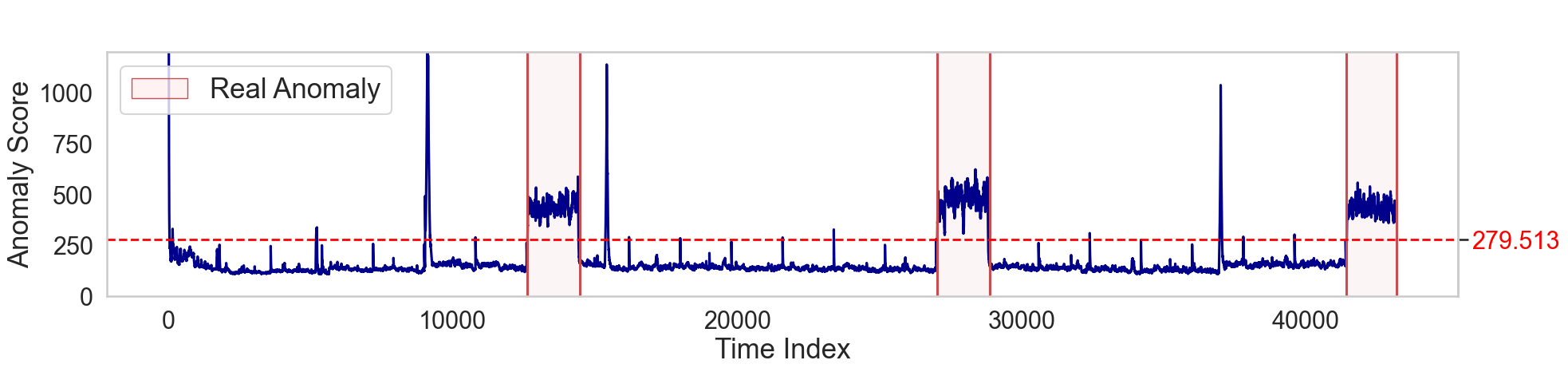}
			\caption{Dense \divadg.}
			\label{fig:t1-dense-divad-g-scores}
		\end{subfigure}
        \begin{subfigure}[b]{1.0\columnwidth}
            \centering
            \includegraphics[width=\columnwidth]{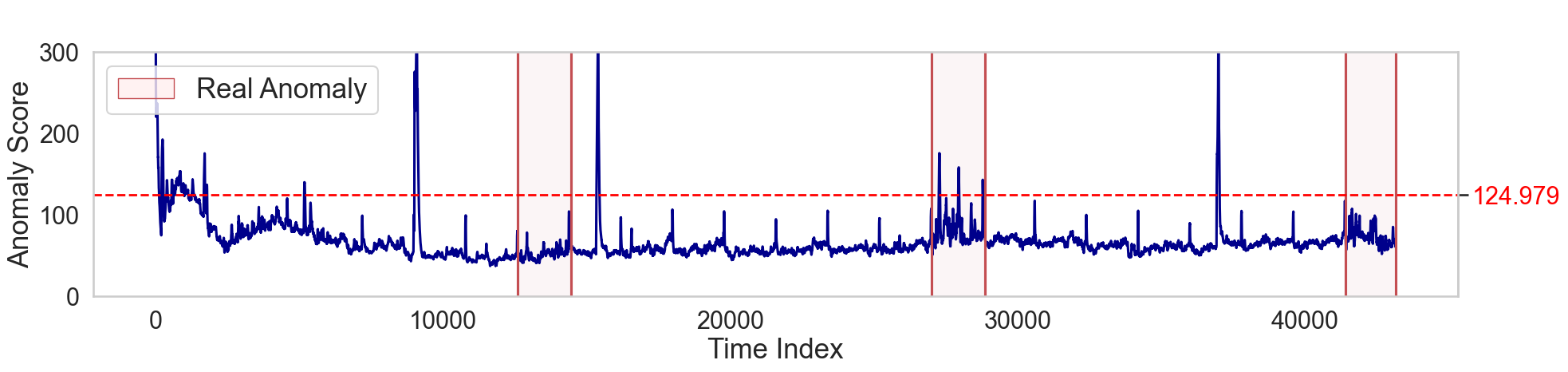}
            \caption{Rec \divadg.}
            \label{fig:t1-rec-divad-g-scores}
        \end{subfigure}
        \caption{Time plots of the anomaly scores of Dense \divadg~and Rec \divadg~for the records in trace \texttt{5\_1\_100000\_63} (Bursty Input), highlighting their peak F1-score thresholds and the ground-truth anomaly ranges.}
        \label{fig:t1-divad-g-scores}
        \Description[Time plots of the anomaly scores of Dense \divadg~and Rec \divadg~for the records in trace 5\_1\_100000\_63 (Bursty Input), highlighting their peak F1-score thresholds and the ground-truth anomaly ranges.]{Time plots of the anomaly scores of Dense \divadg~and Rec \divadg~for the records in trace 5\_1\_100000\_63 (Bursty Input), highlighting their peak F1-score thresholds and the ground-truth anomaly ranges.}
\end{figure}

It is worth noting that the low performance of sequence modeling variants can also be explained by the \emph{unsupervised nature} of our \divad\ methods, resulting in a lack of incentive for them to learn domain-invariant patterns that are sure to preserve anomaly signals. With point modeling, the feature combinations that tended to be domain-invariant were also useful to detect the anomalies of our dataset and setup. For Rec \divadg, however, the \emph{sequential} patterns learned to be shared across domains also tended to be shared between normal data and specific anomaly types. As illustrated in Figure~\ref{fig:rec-divad-g-scores}, showing KDE plots of the anomaly scores assigned by Rec \divadg\ to training normal, test normal and test anomalous records, Rec \divadg\ could accurately perform its DG task, with training and test normal records getting assigned similar anomaly scores (aligned {\color{Green} \textbf{green}} and {\color{NavyBlue} \textbf{blue}} KDEs). This accurate DG however did not result in a better performance, due to Rec \divadg's inability to distinguish some anomalous records from normal data in domain-invariant space (high {\color{NavyBlue} \textbf{blue}} and {\color{Red} \textbf{red}} KDEs overlap). Figure~\ref{fig:t1-divad-g-scores}, showing time plots of the anomaly scores assigned by Dense \divadg\ and Rec \divadg\ in trace \texttt{5\_1\_100000\_63} (Bursty Input), further illustrates this for T1 events specifically. From these figures, we can see that, while Dense \divadg\ accurately deemed T1 records ``more abnormal" than most normal records in the trace, the encoding performed by Rec \divadg\ tended to \emph{remove} most of the anomalous signals from these events.

\section{Details on Time Measurements}
\label{appendix:times}

This section provides details about the time measurements we performed for the VAE and \divad\ variants.

Table~\ref{tab:times} shows the average training and inference step times for the VAE and \divad\ variants on an NVIDIA A100 80GB PCIe, computed across the first 1000 mini-batches of size $B=32$ (skipping the very first ones), with hyperparameters adjusted so as to make \divad\ and VAE directly comparable. Specifically, we ran:

\begin{itemize}
    \item All methods with an encoding dimension of $16$.
    \item Dense VAE, Dense \divadg\ and Dense \divadgm\ using a single hidden layer of $200$ units for the encoder and decoder networks.
    \item Rec VAE, Rec \divadg\ and Rec \divadgm\ using encoders with a 1D convolutional layer of $32$ filters of size $5$ and a stride length of $1$, followed by a GRU layer of $64$ units, and decoders defined symmetrically as per the design of Figure~\ref{fig:conv-rec-ae}.
    \item The anomaly scoring of VAE redefined to consider the variational posterior's \emph{mean} instead of $256$ samples, so as to match \divad's anomaly scoring strategy.
    \item The anomaly scoring of \divad\ based on the class encoding prior $p_{\bsb \lambda}(\mathbf z_y)$. Considering scoring from the aggregated posterior estimate indeed makes negligible differences at inference time, only attributable to the tool used to separately model the distribution. 
\end{itemize}

We set the rest of VAE and \divad's hyperparameters as described in Appendices~\ref{appendix:methods} and \ref{appendix:divad}, respectively. Both VAE and \divad\ were implemented using TensorFlow 2.14.0~\cite{tensorflow} and TensorFlow Probability 0.22.1~\cite{tensorflow-probability}.

\balance
\section{Details on the ASD Experiment}
\label{appendix:asd}

This section provides details about the experiment we conducted on the Application Server Dataset (ASD)~\cite{interfusion}.

We report the performance of TranAD using the implementation of Tuli et al.~\cite{tranad}, and the same preprocessing and hyperparameters as for the Spark streaming dataset. For Rec \divadgm, we consider each server trace as a separate \emph{domain}, and report its performance using the following preprocessing and hyperparameters:

\begin{itemize}
	\item A standardization of the input samples.
	\item Each encoder $\text{NN}_{\bsb \phi_y}$ and $\text{NN}_{\bsb \phi_d}$ with a 1D convolutional layer using $32$ filters of size $5$, a stride length of $1$ and the ReLU activation function, followed by a GRU layer of $32$ units using the $\tanh$ activation function, and a fully-connected layer to output the encoding parameters. 
	\item The decoder $\text{NN}_{\bsb \theta_{yd}}$ defined symmetrically to one encoder as per the design of Figure~\ref{fig:conv-rec-ae}.
	\item A single fully-connected hidden layer of $32$ units for the conditional domain prior network $\text{NN}_{\bsb \theta_{d}}$.
	\item A ReLU activation function followed by a single fully-con\-nected layer of $N_{\text{dom}} = 11$ units (i.e., one per source domain) for the classification head $\text{NN}_{\bsb \omega_{d}}$.
	\item An encoding dimension of $16$, with $K=8$ Gaussian Mixture components.
	\item A KL divergence weight $\beta \in \{1, 5\}$. 
	\item A domain classification weight $\alpha_d = 1,000$. 
    \item A batch size $B=128$.
	\item An anomaly scoring based on the class encoding aggregated posterior estimate $\hat q_{\bsb \phi_y}(\mathbf z_y)$, fitting a Gaussian Mixture distribution with $K=8$ components to the training class encodings.
\end{itemize}

\end{document}